\documentclass[letterpaper]{article} 
\usepackage{aaai24}  
\usepackage{times}  
\usepackage{helvet}  
\usepackage{courier}  
\usepackage[hyphens]{url}  
\usepackage{graphicx} 
\usepackage{natbib}  
\usepackage{caption} 
\usepackage{algorithm}
\usepackage{algorithmic}

\usepackage{newfloat}
\usepackage{listings}
\DeclareCaptionStyle{ruled}{labelfont=normalfont,labelsep=colon,strut=off} 
\floatstyle{ruled}
\newfloat{listing}{tb}{lst}{}
\floatname{listing}{Listing}
\nocopyright 

\usepackage{makecell}
\usepackage{threeparttable}
\usepackage{arydshln}
\usepackage{subfig}
\usepackage{epsfig}
\usepackage{amsmath}
\usepackage{amssymb}
\usepackage{mathbbol}
\usepackage{dsfont}

\usepackage[dvipsnames, svgnames, x11names]{xcolor}
\definecolor{red2}{RGB}{240,191,211}
\definecolor{deepred}{RGB}{219,95,146}
\definecolor{blue2}{RGB}{205,226,247}
\definecolor{deepblue}{RGB}{88,161,230}

\makeatletter
\DeclareRobustCommand\onedot{\futurelet\@let@token\@onedot}
\def\@onedot{\ifx\@let@token.\else.\null\fi\xspace}

\newcommand{\mathcolorbox}[2]{\colorbox{#1}{$ #2$}}

\title{Feature Fusion from Head to Tail for Long-Tailed Visual Recognition}

\author {
    Mengke Li\textsuperscript{\rm 1,2},
    Zhikai Hu\textsuperscript{\rm 3},
    Yang Lu\textsuperscript{\rm 4},
    Weichao Lan\textsuperscript{\rm 3},
    Yiu-ming Cheung\textsuperscript{\rm 3},
    Hui Huang\textsuperscript{\rm 2}\thanks{Corresponding author.}
}
\affiliations {
    \textsuperscript{\rm 1} Guangdong Laboratory of Artificial Intelligence and Digital Economy (SZ), Shenzhen, China \\
    \textsuperscript{\rm 2} College of Computer Science and Software Engineering, Shenzhen University, Shenzhen, China\\
    \textsuperscript{\rm 3} Department of Computer Science, Hong Kong Baptist University, Hong Kong, China\\
    \textsuperscript{\rm 4} Fujian Key Laboratory of Sensing and Computing for Smart City, School of Informatics, Xiamen University, Xiamen, China\\
    limengke@gml.ac.cn, cszkhu@comp.hkbu.edu.hk, luyang@xmu.edu.cn, \{cswclan, ymc\}@comp.hkbu.edu.hk, hhzhiyan@gmail.com
}

\usepackage{bibentry}

\begin{document}

\maketitle

\begin{abstract}
The imbalanced distribution of long-tailed data presents a considerable challenge for deep learning models, as it causes them to prioritize the accurate classification of head classes but largely disregard tail classes.
The biased decision boundary caused by inadequate semantic information in tail classes is one of the key factors contributing to their low recognition accuracy.
To rectify this issue, we propose to augment tail classes by grafting the diverse semantic information from head classes, referred to as head-to-tail fusion (H2T). 
We replace a portion of feature maps from tail classes with those belonging to head classes.
These fused features substantially enhance the diversity of tail classes.
Both theoretical analysis and practical experimentation demonstrate that H2T can contribute to a more optimized solution for the decision boundary.
We seamlessly integrate H2T in the classifier adjustment stage, making it a plug-and-play module. 
Its simplicity and ease of implementation allow for smooth integration with existing long-tailed recognition methods, facilitating a further performance boost. 
Extensive experiments on various long-tailed benchmarks demonstrate the effectiveness of the proposed H2T.
The source code is available at \url{https://github.com/Keke921/H2T}.
\end{abstract}

\section{Introduction}
Deep models have shown remarkable capabilities in diverse visual recognition tasks~\cite{Zhikai2019Triplet, Mengke21FSG, Shervin2022Image, LanWC24pami}, yet their performance heavily relies on data that is evenly distributed across categories. 
In contrast, real-world data typically follows long-tailed distributions~\cite{reed2001pareto, LiuZW19LTOW, zhang2021survey}, which hinders model performance, especially on minority classes (referred to as tail classes in long-tailed data). 
This challenge has become one of the bottlenecks limiting the advancement of deep models.
In recent times, a plethora of methods have emerged to address the problem of severe class imbalance in long-tailed data from different aspects, such as class-balancing methods~\cite{Nitesh2002SMOTE, Huang2016CVPR, Mengye2018Learning, cui2019class, Tsung2020Focal, HuangC2020Deep, Hong2021CVPR, ParkS2021ICCV}, re-margining methods~\cite{Kaidi2019, LiMK2022GCL, LiMK2022KPS}, data augmentation~\cite{kim2020m2m, Wang2021RSG, Li2021MetaSAug, ParkS2022Majority}, and ensembling learning~\cite{WangXD21RIDE, LiBL2022Trustworthy, Jin2023shike}.
While these methods demonstrated the capability to yield robust predictions, they primarily focus on training a new model to acquire a relatively balanced embedding space and/or assembling multiple diverse networks. 
However, they overlook acquiring a more optimal classifier (determining the decision boundary), which is crucial for fully releasing the potential of the acquired backbone.

\begin{figure}
    \centering
    \includegraphics[width=0.9\linewidth]{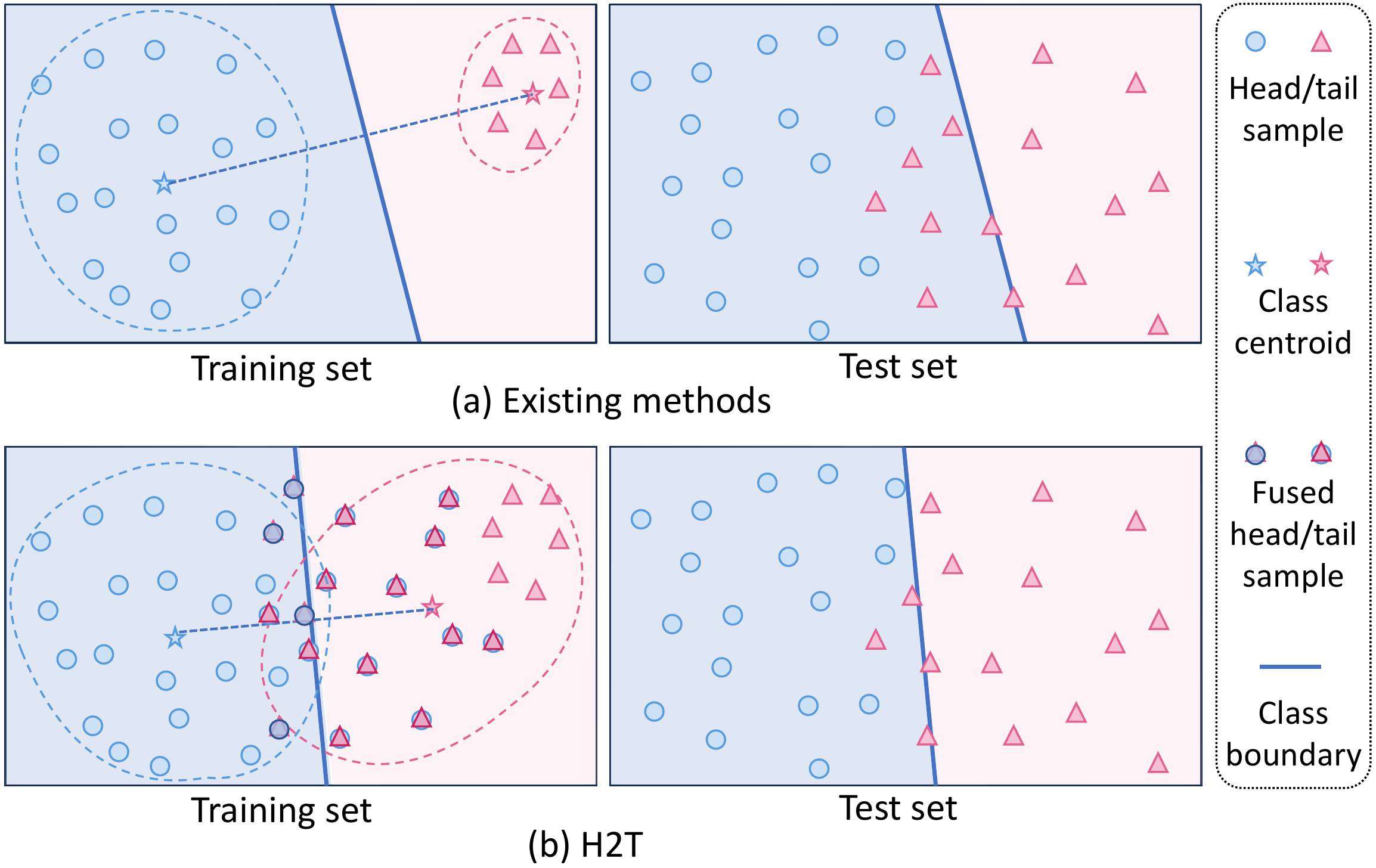}  
    \caption{Comparison between decision boundaries produced by (a) existing methods and (b) proposed H2T.}
\label{fig:dec_bnd}
\end{figure}

This paper thereby aims to obtain a more effective classifier with the trained backbone.
It is widely recognized that increasing class margins~\cite{Kaidi2019, Deng2019ArcFace} and/or tightening the intra-class space~\cite{wang2018cosface,Wang2017NormFace} for the training set can improve the generalization performance of the model~\cite{Zhikai2021MTFH, HuZK23Joint}.
However, while the clear margin is effective for balanced data, it may still have a biased decision boundary for long-tailed data.
Take the binary case in the embedding space of the obtained backbone as an example, the decision boundary is usually the midline connecting the two class centroids. 
The head class has sufficiently sampled and its embedding space is fully occupied~\cite{xiao2021does}. 
On the opposite, the tail class suffers from a scarcity of samples, leading to sparsely distributed semantic regions. 
The bias in the tail centroid persists due to head squeeze even if the existence of a clear margin. 
As a result, during the inference stage, a considerable number of samples that differ from the training set emerge, causing the erosion of well-defined margins.
Consequently, numerous tail class samples are misclassified as head classes, as illustrated in Figure~\ref{fig:dec_bnd}(a).

To enrich the sparse tail class semantics and calibrate the bias in tail classes, we propose a direct and effective solution, named head-to-tail fusion (H2T), which grafts partial semantics from the head class on the tail class. 
In particular, the fusion operation in H2T involves the direct replacement of certain features of tail samples with those of head samples.
This is based on the assumption that predictions on tail classes that have rare instances are easily affected by head classes that appear frequently.
Transferring the head semantics can effectively fill the tail semantic area and the category overlap, which compels the decision boundary to shift closer to a more optimal one, as shown in Figure~\ref{fig:dec_bnd}(b). 
Ultimately, the generalization performance of the tail classes can be improved. 
To streamline the fusion operation, we design an easy-to-implement strategy that takes full use of the obtained features without retraining the backbone. 
Specifically, we randomly substitute several channels in the feature maps of the class-balanced data with those of the head-biased data. 
This makes the feature maps of tail classes with a high probability of being fused with head classes.
We apply H2T to the classifier tuning stage, and with just a few lines of code and a few epochs of training, remarkable performance improvements can be achieved.
In addition, H2T does not alter the structure of the backbone or increase the network parameters.
This characteristic makes it highly adaptable and compatible with various existing techniques, facilitating its seamless incorporation into diverse existing methods.
The main contributions of this paper are summarized as follows:
\begin{itemize}
    \item We propose a novel H2T that borrows information from head classes and applies it to augment tail classes without additional data or network parameters, which can release the potential of the well-trained backbone and obtain better decision boundaries.
    \item We devise a simple fusion strategy that partially substitutes the feature maps of an instance-wise sampling branch with those of a class-balanced sampling branch. 
    This approach enables H2T to be implemented with ease and without requiring any modifications to the original backbone structure.     
    \item Considerable performance improvement is achieved in both single and multi-expert models by integrating the H2T module into the existing structures. 
    Extensive experiments on popular benchmarks validate the efficacy of the proposed method. 
\end{itemize}

\section{Related Work}
\label{sec:related_work}
This section mainly makes an overview of methods based on data augmentation which is the most relevant regime.

\noindent\textbf{Classical Augmentation:}
Classical augmentation methods including flip, rotate, crop, padding, and color jittering etc.~\cite{Szegedy2015Going,he2016deep} have been widely applied in deep models.
On long-tailed data, these methods can improve model robustness and prevent overfitting to specific input patterns to a certain extent. 
Recently, AutoAugment~\cite{CubukED19AutoAugment}, Fast AutoAugment~\cite{LimS19FastAA}, Population based augmentation~\cite{HoD19PBA} and Randaugment~\cite{CubukED20Randaugment}, etc. have been proposed to determine the best data augmentation strategy for a dataset automatically. 
These methods have been proven to be effective in increasing the classification accuracy for DNNs and also have demonstrated their efficacy on long-tailed data \cite{RenJW2020Balanced, CuiJQ2021parametric, liJ2022nested}.

\noindent\textbf{MixUp Based Augmentation:}
MixUp~\cite{Hongyi2018} is an effective data augmentation method that linearly combines a pair of row images and their labels. 
This method adds complexity control to the uncovered space in the data space through linearly interpolating discrete sample points, which reduces the model generalization error. 
Numerous studies~\cite{zhang2021bag, mislas21, LiMK2022GCL} have experimentally proven that MixUp can significantly improve long-tailed recognition performance.
CutMix~\cite{YunSD2019cutmix}, a variant of MixUp, augments data by cutting a patch from one input image and pasting it into another image within the training set. 
The ground truth labels are combined in proportion to the patch areas. 
Park~et~al.~\cite{ParkS2022Majority} propose mixing the foreground patches from tail classes with the background images from head classes via CutMix, enabling the transfer of context-rich background information from head to tail. 

\noindent\textbf{New Images Based Augmentation:}
In addition to the aforementioned methods, researchers have proposed alternative approaches that generate new informative samples for tail classes. 
One such method is M2m~\cite{kim2020m2m}, which translates samples from head classes to tail classes using a pretrained auxiliary classifier.
Wang~et~al.~\cite{Wang2021RSG} utilize the learned encoded variation information from head classes to synthesize images for tail classes. 
Zada~et~al.~\cite{Zada22noise} directly exploit pure noise images as tail class samples. 
Zhang~et~al.~\cite{zhang2021bag} utilize class activation maps (CAM)~\cite{Zhou2016Learning} to separate image foreground and background, followed by augmenting the foreground by flipping, rotating, and scaling, etc. The new samples are then generated by overlaying the augmented foreground on the unchanged background.   

\noindent\textbf{Feature Level Augmentation:}
Data augmentation can also be performed in feature space. 
For example, Manifold MixUp~\cite{verma2019manifold} conducts the linear interpolation on the features of input images, which has been shown to yield better results in long-tail learning~\cite{zhang2021bag, bbn20}. 
Chu~et~al.~\cite{Peng2020Feature} use CAM to decompose features into class-generic and class-specific components. Tail classes are then augmented by combining their class-specific components with the class-generic components.
Another promising technique is to transfer knowledge from head classes to tail classes. 
For example, Yin~et~al.~\cite{yin2019feature} and Liu~et~al.~\cite{Liu2020Deep} enrich tail classes with the intra-class variance of head classes to balance the class distributions in feature space. 
Recently, meta-learning-based augmentation has been leveraged to address class imbalance.
Liu~et~al.~\cite{Liu2022Memory} design a meta-embedding that uses a memory bank to enrich tail classes. 
MetaSAug~\cite{Li2021MetaSAug} learns class-wise covariance matrices by minimizing loss on a small balanced validation set and subsequently uses the learned matrices to semantically augment tail classes.

\section{Methodology}\label{sec:method}
\subsection{Preliminaries}
We formally define the basic notations used in this paper before going into detail about our proposed method.
We use $\{x,y\}$ to represent one input image and its corresponding label. The total number of classes is denoted by $C$, thus, we have $y \in \{0, 1, \cdots, C-1\}$. 
The training set includes $N$ samples. 
Suppose that class $i$ has $n_i$ training samples. 
Then $N=\sum_i n_i$. 
For simplicity, we suppose $n_0 \geq n_1 \geq \cdots \geq n_{C-1}$. 
Feed $x$ into the backbone, we can obtain its feature maps before the last pooling layer, denoted as $\mathcal{F} = \left[F_{0}, F_{1}, \cdots, F_{d-1} \right] \in \mathbb{R}^{w_{F} \times h_{F} \times d}$, where $w_{F}$ and $ h_{F}$ represent the width and height of the feature map, respectively.
$d$ is the dimension of features in embedding space. 
The representation after the last pooling layer is ${f} \in \mathbb{R}^{d} $.
The weight of the linear classifier is represented as $\textbf{W}=[w_0, w_1, \cdots, w_{C-1}] \in \mathbb{R}^{d \times C} $, where $w_i$ represents the classifier weight for class $i$. 
We use the subscripts $h, m$ and $t$ to indicate head, medium and tail classes, respectively. 
$z_i=w_i^Tf$ represents the predicted logit of class $i$, where the subscript $i=y$ denotes the target logit and $i\neq y$ denotes the non-target logit.

\subsection{Motivation} \label{sec:motivation}

\begin{figure*}
    \centering
    \includegraphics[width=\linewidth]{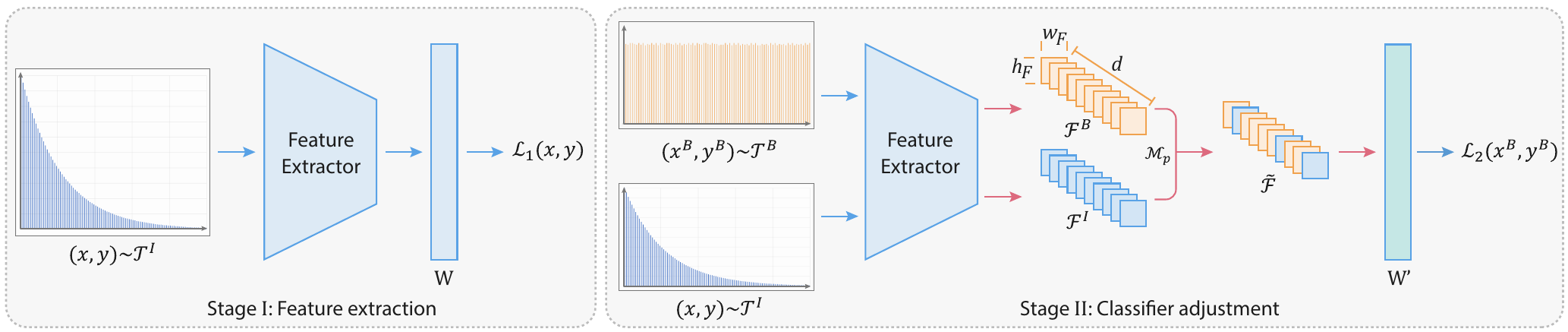}  
    \caption{Framework of head-to-tail fusion (H2T).}
\label{fig:framework}
\end{figure*}

Head classes are significantly more numerous than tail classes in long-tailed data, resulting in a biased classifier, which leads to poor performance on the test set of tail classes. 
The straightforward solution involves increasing the importance or frequencies of tail classes ~\cite{Kaidi2019, bbn20,shin2017deep}.
They can increase the performance of tail classes, nevertheless, they also entail an elevated risk of overfitting. 
To adjust the decision boundary and prevent overfitting, it is crucial to enhance the diversity of tail class samples~\cite{zhang2021survey, yang2022survey}. 
Unfortunately, directly obtaining more samples is generally infeasible.
We can consider enriching the tail class by maximizing the utilization of existing features.
Typically, the misclassified samples are unseen instances from the training set, and these instances tend to be interspersed in the vicinity of the decision boundary. 
This pattern provides us an opportunity to augment the diversity within tail classes and populate class margins with semantically meaningful samples. 
This augmentation process contributes to an optimal decision boundary.
To achieve this, we simulate the potential unseen samples by directly borrowing semantic information from head classes to augment tail classes. 
By doing so, the rich semantic information can be transferred from well-represented head classes to tail classes.


\subsection{Methodology: Fusing Head Features to Tail}\label{sec:H2T_method}
We fuse the features of head classes to the tail to exploit the abundant closet semantic information.
This operation can enrich the tail classes and expand their embedding space distribution. 
The fusion process is formulated as:
\begin{equation}\label{eq:fh2t}
    \tilde{\mathcal{F}} = \mathcal{M}_{p} \otimes \mathcal{F}_t  + \overline{\mathcal{M}}_{p}\otimes \mathcal{F}_h ,
\end{equation}
where $\mathcal{M}_{p}$ is the mask stacked with multiple 1 matrices $\mathbf{I}_M= \mathbb{1}^{w_{F} \times h_{F}}$ and 0 matrices $\mathbf{O}_M = \mathbb{0}^{w_{F} \times h_{F}}$. The total number of all the 1 and 0 matrices is $d$. 
$\overline{\mathcal{M}}{p}$ is the complement of $\mathcal{M}{p}$, that is, the indices of 0 matrices in $\overline{\mathcal{M}}{p}$ correspond to the 1 matrix in $\mathcal{M}{p}$, and vice versa.
The subscript $p$ of the mask matrices represents the fusion ratio. Specially, the number of $\mathbf{I}_M$ and $\mathbf{O}_M$ is $[d \times p]$ and $d- [d\times p]$, respectively. 
$[\cdot]$ means rounding operation.
$\tilde{\mathcal{F}}$ is then passed through a pooling layer and classifier to predict the corresponding logits $\mathbf{z} = [z_0, z_1, \cdots, z_{C-1}]$. 
Different loss functions, such as CE loss, MisLAS~\cite{mislas21}, or GCL~\cite{LiMK2022GCL}, to name a few, can be adopted. 
The backbone $\phi$ can be the single model~\cite{he2016deep} as well as the multi-expert model~\cite{WangXD21RIDE, XiangLY2020LFME}. 
We exploit the two-stage training~\cite{decouple20} and apply H2T in stage II.

\begin{algorithm}[t]
\renewcommand{\algorithmicrequire}{\textbf{Input:}}
\renewcommand{\algorithmicensure}{\textbf{Output:}}
\caption{H2T}\label{alg:h2t}
\begin{algorithmic}[1]
\REQUIRE {Training set, fusion ration $p$ \;}
\ENSURE {Trained model\;}
\STATE Initialize the model $\phi$\ randomly 
\FOR {$iter=1$ to $E_0$}
\STATE Sampling batches of data $(x,y)\sim \mathcal{T}^I$ from the instance-wise sampling data
\STATE Obtain the feature map $\mathcal{F} = \phi_{\theta}(x)$ 
\STATE Calculate logits $z = \mathbf{W}^T \mathbf{f}$ and loss $\mathcal{L}_1(x,y)$\;
\STATE $\phi = \phi - \alpha \nabla_{\phi} \mathcal{L}_1((x,y);\phi)$.
\ENDFOR
\FOR {$iter=E_0+1$ to $E_2$}
\STATE Sample batches of data $(x^B,y^B)\sim \mathcal{T}^B$ and $(x,y)\sim \mathcal{T}^I$\;
 {\color{deepred} \STATE Obtain feature maps $\mathcal{F}^B = \phi_{\theta}(x^B)$ and $\mathcal{F}^I = \phi_{\theta}(x)$ \;
\STATE Fuse feature maps by Eq.~\ref{eq:fh2t} to obtain $\tilde{\mathcal{F}}$ \; 
\STATE Input $\tilde{\mathcal{F}}$ to pooling layer to obtain $\tilde{\mathbf{f}}$, then calculate the logits by $\tilde{z} = \mathbf{W}^T \tilde{\mathbf{f}}$ and the loss by $\mathcal{L}_2(x^B,y^B)$ \;}
\STATE Froze the parameters of representation learning $\phi^r$, and finetune the classifier parameters $\phi^c$:
$\phi^c = \phi^c- \alpha \nabla_{\phi^c} \mathcal{L}_2((x^B,y^B);\phi^c)$.
\ENDFOR
\end{algorithmic}  
\end{algorithm}

The selection of features to be fused poses a tedious task during training since visual recognition tasks often encompass a vast number of categories, and it cannot be guaranteed that each minibatch contains the required categories. 
We thereby devise a simple strategy to facilitate an easy-to-execute fusion process.
This strategy involves sampling two versions of data: 1) class-balanced data $\mathcal{T}^B$ used to balance the empirical/structural risk minimization (ERM/SRM) of each class, which is fed into the fused branch, and 2) instance-wise data $\mathcal{T}^I$ has a high probability to obtain head class samples, which is fed into the fusing branch. 
The sampling rates $p^B_i$ and $p^I_i$ for class $i$ within the sets $\mathcal{T}^B$ and $\mathcal{T}^I$, respectively, are calculated by
\begin{equation}\label{eq:sam_rate}
     p^B_i = \frac{1}{C}, \:
     p^I_i = \frac{n_i}{N}.
\end{equation}
Balanced sampling data ensures that each class is sampled with equal probability $\frac{1}{C}$. 
The fewer the number of samples, the higher the probability of being resampled multiple times.
The instance-wise sampling branch deals with the original data distribution, resulting in head-biased data. 
Therefore, the head classes have a higher probability of being sampled.
Next, we can use the feature maps $\mathcal{F}^B$ and $\mathcal{F}^I$ obtained from $\mathcal{T}^B$ and $\mathcal{T}^I$ to substitute $\mathcal{F}_t$ and $\mathcal{F}_h$ in Eq.~(\ref{eq:fh2t}). 
By doing so, the features of the repeatedly sampled tail classes will be fused with the head class features with a higher probability. 
In contrast to the common practice of linearly combining a pair of inputs and their labels for augmentation, we only use the labels of the fused branch, namely the balance-sampled data, as the ground truth.
By adopting this method, the semantics of head classes are utilized to enrich tail classes.
The proposed framework is illustrated in Figure~\ref{fig:framework} and Algorithm~\ref{alg:h2t}. 
The key steps are highlighted in {\color{deepred}rose-red}.

\subsection{Rationale Analysis}\label{sec:ana}
Although the proposed H2T appears intuitive and straightforward on the surface, it is built on a foundation of rationale. 
We explore its theoretical rationality in depth in this section.
For ease of analysis, we assume that the feature maps to be fused are rearranged in order without loss of generality. 
Then, after the pooling layer, the feature can be written as
$f_i^T = [ \dot{f}_i^T,\ddot{f}_i^T ]$, where $\dot{f}_{i}$ and $\ddot{f}_{i}$ denote the portions of features that are retained and to be fused, respectively.
$w_i^T = [\dot{w}_i^T, \ddot{w}_i^T]$ is the corresponding classifier weights.

\begin{figure}[tb]
\centering
\includegraphics[width=0.95\linewidth]{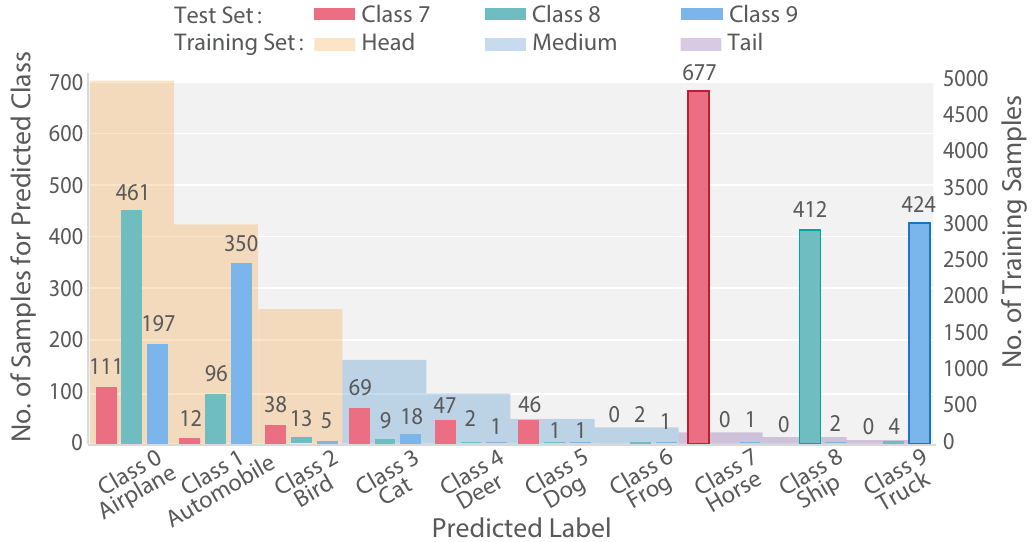}
\caption{Frequency of predictive labels for tail classes samples. A large number of tail class samples are incorrectly recognized as head classes.}
\label{fig:dist}
\end{figure}

On the one hand, we expect $z_t > z_h$ for tail class, so that
\begin{equation}\label{eq:correct_z}
    \begin{split}
         w_t^T f_t & > w_h^T f_t  \:\: \Rightarrow \\
         \dot{w}_{t}^T\dot{f}_{t}+\ddot{w}_{t}^T\ddot{f}_{t}  & > \dot{w}_{h}^T\dot{f}_{t}+\ddot{w}_{h}^T\ddot{f}_{t}  .
    \end{split}
\end{equation}
However, as shown in Figure~\ref{fig:dist}, numerous tail class samples are incorrectly classified into head classes. 
Therefore, the trained model actually predicts $z_h > z_t$, namely,
\begin{equation}\label{eq:actual_z}
    \dot{w}_{t}^T\dot{f}_{t}+\ddot{w}_{t}^T\ddot{f}_{t} < \dot{w}_{h}^T\dot{f}_{t}+\ddot{w}_{h}^T\ddot{f}_{t}.
\end{equation}

\begin{figure}[tb]
    \centering
    \includegraphics[width=\linewidth]{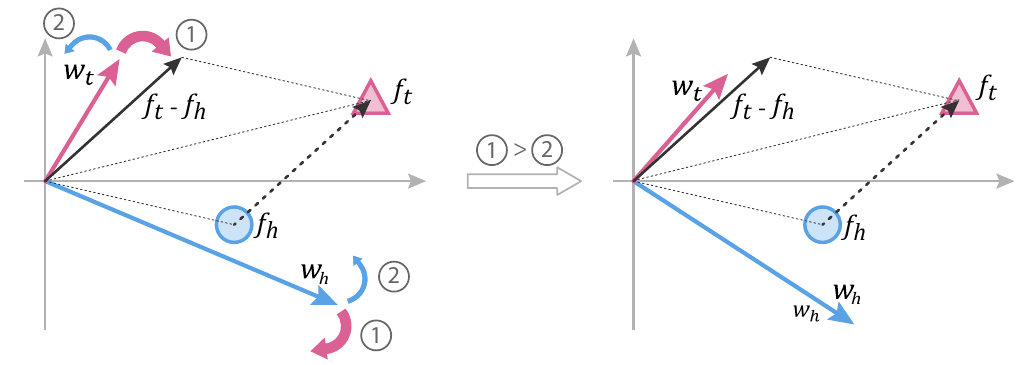}
    \caption{Rationale analysis of H2T. Forces {\footnotesize\textcircled{\scriptsize{1}}} and {\footnotesize\textcircled{\scriptsize{2}}} are generated by Eq.~(\ref{eq:theta2}) and Eq.~(\ref{eq:theta}), respectively. {\footnotesize\textcircled{\scriptsize{1}}} $>$ {\footnotesize\textcircled{\scriptsize{2}}} makes the tail sample to ``pull" closer to $w_t$ and ``push" further away from $w_h$, leading to the adjustment of decision boundary and enlargement of the tail class space.}
\label{fig:rationale}
\end{figure}

After fusing the head to the tail in stage II, the feature of tail class is $\tilde{f}_t = [\dot{f}_{t}, \ddot{f}_{h}]$, and the corresponding logit is $\tilde{z}_t$.
Our training goal is still to make the target logit larger than the non-target class, namely, $\tilde{z}_t > \tilde{z}_h$. 
Therefore, we have
\begin{equation}\label{eq:fuse_z}
    \begin{split}
    w_t^T \tilde{f}_t &> w_h^T \tilde{f}_t \Rightarrow      \\
    \begin{bmatrix} \dot{w}_t \\ \ddot{w}_t \end{bmatrix}^T 
    \begin{bmatrix} \dot{f}_t\\ \ddot{f}_h \end{bmatrix} & > 
    \begin{bmatrix} \dot{w}_h\\ \ddot{w}_h \end{bmatrix}^T 
    \begin{bmatrix} \dot{f}_t\\ \ddot{f}_h \end{bmatrix} \Rightarrow \\
    \dot{w}_{t}^T\dot{f}_{t}+\ddot{w}_{t}^T\ddot{f}_{h} & > \dot{w}_{h}^T\dot{f}_{t}+\ddot{w}_{h}^T\ddot{f}_{h}.
    \end{split}    
\end{equation}
Adding the last line in Eq.~(\ref{eq:fuse_z}) to Eq.~(\ref{eq:actual_z}), we can obtain that
\begin{equation}\label{eq:fuse1}
    \ddot{w}_h^T (\ddot{f}_t-\ddot{f}_h) > \ddot{w}_t^T (\ddot{f}_t-\ddot{f}_h).
\end{equation}
We use $\ddot{\theta}_*$ ($*$ is $h$ or $t$) to represent the angle between $\ddot{w}_*$ and $\ddot{f}_t-\ddot{f}_h$. Eq.~(\ref{eq:fuse1}) is further simplified to
\begin{equation}\label{eq:theta}
    |\ddot{w}_h| \cos\ddot{\theta}_h > |\ddot{w}_t| \cos\ddot{\theta}_t.
\end{equation}
On the other hand, similar to Eq.~(\ref{eq:correct_z}), for logits of head classes, we have
\begin{equation}\label{eq:correct_head}
        \dot{w}_{h}^T\dot{f}_{h}+\ddot{w}_{h}^T\ddot{f}_{h} > \dot{w}_{t}^T\dot{f}_{h}+\ddot{w}_{t}^T\ddot{f}_{h}.
\end{equation}
Add the last line in Eq.~(\ref{eq:fuse_z}) to Eq.~(\ref{eq:correct_head}), we obtain that
\begin{equation}\label{eq:theta2}
    |\dot{w}_t| \cos\dot{\theta}_t > |\dot{w}_h| \cos\dot{\theta}_h.
\end{equation}
Eq.~(\ref{eq:theta}) forces $w_h$ to move closer to tail class samples while pushing $w_t$ further away from them, thus increasing the distribution span of head classes.
Conversely, Eq.~(\ref{eq:theta2}) exerts an opposite force. 
Figure~\ref{fig:rationale} geometrically interprets the rationale of H2T.
Notably, when the fusion ratio $p$ is small, Eq.~(\ref{eq:theta2}) exerts a greater force, enabling the classifier to be well-calibrated. 
As $p$ grows, the item $\ddot{w}_t^T\ddot{f}_h > \ddot{w}_h^T\ddot{f}_h$ in Eq.~(\ref{eq:fuse1}) progressively becomes more dominant. 
Under this case, even though Eq.~(\ref{eq:theta}) encourages $w_t$ to move away from tail class samples, which is larger than the force of Eq.~(\ref{eq:theta2}), $\ddot{w}_t^T\ddot{f}_h > \ddot{w}_h^T\ddot{f}_h$ attracts $w_t$ towards head class samples, thus broadens the distribution span of tail classes. 
Consequently, the classifier increases its performance in the tail class, regardless of the value of $p$. 
We also experimentally validate this analysis Section~\ref{sec:ablation}. 

\section{Experiments}

\subsection{Datasets and Metrics}
We evaluate our proposed H2T on four widely-used benchmarks: CIFAR100-LT~\cite{Kaidi2019}, ImageNet-LT~\cite{LiuZW19LTOW}, iNaturalist 2018~\cite{Horn_2018_CVPR}, and Places-LT~\cite{LiuZW19LTOW}. 
CIFAR100-LT is a small-scale dataset that is sub-sampled from the balanced version CIFAR100~\cite{krizhevsky2009learning}.
We used the imbalance ratios ($\rho=\frac{n_{max}}{n_{min}}$) 200, 100, and 50~\cite{Peng2020Feature}. 
The original versions of imageNet-2012~\cite{ILSVRC15} and Places 365~\cite{zhou2017places} are also balanced datasets. 
We use the same settings as Liu~et~al.~\cite{LiuZW19LTOW} to obtain the long-tailed versions. 
iNaturalist is collected from all over the world and is naturally heavily imbalanced. 
The 2018 version~\cite{Horn_2018_CVPR} is utilized in our experiment.
We also report the comparison results on CIFAR10-LT~\cite{Kaidi2019} in the Appendix. 
As for the metrics, besides top-1 classification accuracy, following Liu~et~al.~\cite{LiuZW19LTOW}, the accuracy on three partitions: head ($n_i > 100$), medium ($20<n_i\leq 100$) and tail ($n_i\leq 20$) on large-scale datasets are also compared.  

\subsection{Basic Settings}
SGD with a momentum of 0.9 is adopted for all datasets. Stage II is trained with 10 epochs.
For CIFAR100-LT, we refer to the settings in Cao~et~al.~\cite{Kaidi2019} and Zhong~et~al.~\cite{mislas21}. 
The backbone network is ResNet-32~\cite{he2016deep}. 
Stage I trains for 200 epochs. 
The initial learning rate is 0.1 and is decayed at the $160^{th}$ and $180^{th}$ epochs by 0.1.  
The batch size is 128. 
For imageNet-LT and iNaturalist 2018, we use the commonly used ResNet-50. 
For Places-LT, we utilize the ResNet-152 pre-trained on imageNet.
We reproduce the prior methods that provide the hyper-parameters for a fair comparison. 
For those methods that do not provide hyperparameters or official codes, we directly report their results presented in the original paper.

\subsection{Comparisons to Existing Methods} 
\label{sec:com}

\textbf{Compared Methods:}\label{sec:com_method}
For single model, we compare the proposed H2T with the following three kinds of methods:
\textit{two-stage methods}, i.e., LDAM-DRW~\cite{Kaidi2019}, decoupling representation (DR)~\cite{decouple20}\footnote{We reproduce DR with MixUp for a fair comparison.}, MisLAS~\cite{zhang2021bag}, and GCL~\cite{LiMK2022GCL};
\textit{decision boundary adjustment method}, i.e., Adaptive Bias Loss (ABL)~\cite{jin2023optimal};
\textit{data augmentation methods}, i.e., MBJ~\cite{Liu2022Memory}, and CMO~\cite{ParkS2022Majority}. 
We report the results of CE loss and balanced softmax cross-entropy loss (BSCE)~\cite{RenJW2020Balanced} with CMO (abbreviated as CE+CMO and BSCE+CMO, respectively). 
The results of H2T accompanied by DR (CE loss is utilized), MisLAS, and GCL are reported.
For multi-expert models, we compare BBN~\cite{bbn20}, RIDE~\cite{WangXD21RIDE}, ACE~\cite{Cai2021ACE}, and ResLT~\cite{Cui2022reslt}. 
Both RIDE and ResLT use 3-expert architectures. 
For Places-LT, we report the results of RIDE with linear and cosine classifiers.

\begin{table}[tb]
 \centering  
 \caption{Comparison results on CIFAR100-LT.} %
 \label{tab:cifar_results}
 \resizebox{0.9\linewidth}{!}
  {\begin{threeparttable}
  \small
  \begin{tabular}{|l| c c c|}
  \hline 
  \multicolumn{1}{|c|}{Imbalance Ratio}  &200 &100 &50 \\
  \hline 
  \multicolumn{4}{|c|}{Single Model} \\
  \hline 
  CE loss &35.99 & 39.55 &45.40\\
  LDAM-DRW \cite{Kaidi2019} &38.91 &42.04 &47.62\\
  DR+MU~\cite{decouple20} & 41.73 & 45.68 & 50.86\\
  MisLAS~\cite{mislas21} &43.45 &46.71 &52.31 \\
  MBJ$^\star$~\cite{Liu2022Memory} & - & 45.80 & 52.60 \\
  GCL~\cite{LiMK2022GCL} & \textbf{44.76} & 48.61 & \textbf{53.55} \\
  CE+CMO$^\star$~\cite{ParkS2022Majority} & - & 43.90 & 48.30 \\
  BSCE+CMO$^\star$~\cite{ParkS2022Majority}& - & 46.60 & 51.40 \\
  ABL$^\star$~\cite{jin2023optimal}& - & 46.80 & 52.10 \\
  \hdashline 
  {DR+H2T} & 43.95 & \textbf{47.73}&  52.95 \\
  {MisLAS+H2T} & 43.84  & 47.62  & 52.73 \\
  {GCL+H2T} & \underline{\textbf{45.24}} & \underline{\textbf{48.88}}  & \underline{\textbf{53.76}} \\ 
  \hline
    \multicolumn{4}{|c|}{Multi-Expert Model} \\
  \hline
  BBN~\cite{bbn20}  & 37.21 & 42.56 & 47.02 \\
  RIDE~\cite{WangXD21RIDE}  & 45.84 & 50.37 & \textbf{54.99} \\
  ACE$^\star$~\cite{Cai2021ACE}  & - & 49.40 & 50.70 \\
  RIDE+CMO$^\star$~\cite{ParkS2022Majority}  &- & \textbf{50.00} & 53.00\\
  ResLT~\cite{Cui2022reslt} & 44.26 & 48.73 & 53.81 \\
  \hdashline 
  RIDE+H2T & \underline{\textbf{46.64}}  & \underline{\textbf{51.38}} &\underline{\textbf{55.54}} \\
  ResLT+H2T  & \textbf{46.18 } & 49.60  & 54.39 \\
  \hline
 \end{tabular}
    \begin{tablenotes}
    \footnotesize
    \item \textbf{Note}: 
    The backbone is ResNet-32.  $\star$ denotes the results quoted from the corresponding papers.
    The best and the second-best results are shown in \underline{\textbf{underline bold}} and \textbf{bold}, respectively. 
    \end{tablenotes}
 \end{threeparttable}
 }
\end{table}

\begin{table}[tb]
\renewcommand{\thefootnote}{\fnsymbol{footnote}}
 \centering  
 \caption{Comparison results on imageNet-LT.} \label{tab:img} 
 \resizebox{0.93\linewidth}{!}
  {\begin{threeparttable}
  \begin{tabular}{|l |c c c |c|}  
  \hline
  \multicolumn{1}{|c|}{Method}  &Head &Med &Tail & All \\
  \hline 
  \multicolumn{5}{|c|}{Single Model} \\
  \hline 
  CE loss & 64.91 & 38.10  &11.28  &44.51 \\
  LDAM-DRW \cite{Kaidi2019}  & 58.63  &48.95  & 30.37 & 49.96 \\
  DR~\cite{decouple20}  & 62.93 & 49.77 & 33.26 & 52.18 \\ 
  MisLAS~\cite{mislas21} & 62.53 & 49.82 & 34.74 & 52.29 \\ 
  MBJ$^\star$~\cite{Liu2022Memory}  & 61.60 & 48.40 & 39.00 & 52.10  \\
  GCL~\cite{LiMK2022GCL} & 62.24 & 48.62 & 52.12 & \textbf{54.51} \\ 
  CE+CMO$^\star$~\cite{ParkS2022Majority} & 67.00 & 42.30 & 20.50 & 49.10  \\
  BSCE+CMO$^\star$~\cite{ParkS2022Majority} & 62.00 & 49.10 & 36.70 & 52.30 \\
  ABL$^\star$~\cite{jin2023optimal} & 62.60 & 50.30 & 36.90 & 53.20 \\ 
  \hdashline 
  DR+H2T  & 63.26 & 50.43 & 34.11 & 52.74\\
  MisLAS+H2T & 62.42 & 51.07 & 35.36 &  52.90  \\ 
  GCL+H2T & 62.36 & 48.75 & 52.15 & \underline{\textbf{54.62}} \\  
  \hline
    \multicolumn{5}{|c|}{Multi-Expert Model} \\
  \hline
  BBN~\cite{bbn20}  & - & - & - & 48.30  \\
  RIDE~\cite{WangXD21RIDE} & 69.59 & 53.06  & 30.09 & 55.72 \\ 
  ACE$^\star$ ~\cite{Cai2021ACE} & - & - & - & 54.70 \\
  RIDE+CMO$^\star$~\cite{ParkS2022Majority}  & 66.40 & 53.90 & 35.60 & \textbf{56.20}\\
  ResLT~\cite{Cui2022reslt}  & 59.39 & 50.97 & 41.29 & 52.66 \\ 
  \hdashline 
  RIDE+H2T & 67.55 & 54.95 & 37.08 & \underline{\textbf{56.92}}  \\ 
  ResLT+H2T  & 62.29 & 52.29 & 35.31 & 53.39 \\  
  \hline
 \end{tabular}
    \begin{tablenotes}
    \item \textbf{Note}: 
     The backbone is ResNet-50. Others are the same as Table~\ref{tab:cifar_results}.
    \end{tablenotes}
 \end{threeparttable}
}
\end{table}

\begin{table}[tb]
\renewcommand{\thefootnote}{\fnsymbol{footnote}}
 \centering  
 \caption{Comparison results on iNaturalist 2018.} \label{tab:iNat} 
 \resizebox{0.93\linewidth}{!}
  {\begin{threeparttable}
  \begin{tabular}{|l |c c c |c|}
  \hline 
  \multicolumn{1}{|c|}{Method} &Head &Med &Tail & All \\
  \hline 
  \multicolumn{5}{|c|}{Single Model} \\
  \hline 
  CE loss & 76.10  & 69.05  & 62.44  & 66.86 \\
  LDAM-DRW \cite{Kaidi2019} & - &- &- & 68.15 \\
  DR~\cite{decouple20} & 72.88 & 71.15 & 69.24 & 70.49\\  
  MisLAS~\cite{mislas21}   & 72.52 & 72.08 & 70.76 & 71.54  \\
  MBJ$^\star$~\cite{Liu2022Memory} & - & - & - & 70.00  \\
  GCL~\cite{LiMK2022GCL}  & 66.43 & 71.66 & 72.47 & 71.47 \\ 
  CE+CMO$^\star$~\cite{ParkS2022Majority} & 76.90 & 69.30 & 66.60 & 68.90 \\     
  BSCE+CMO$^\star$~\cite{ParkS2022Majority} & 68.80 & 70.00 & 72.30 & 70.90 \\  
  ABL$^\star$~\cite{jin2023optimal} & - & - & - & 71.60 \\  
  \hdashline 
  DR+H2T  & 71.73 & 72.32 & 71.30 & \textbf{71.81}  \\ 
  MisLAS+H2T  & 69.68 & 72.49 & 72.15 & \underline{\textbf{72.05}} \\       
  GCL+H2T &67.74 & 71.92 & 72.22 & 71.62 \\
  \hline
    \multicolumn{5}{|c|}{Multi-Expert Model} \\
  \hline
  BBN~\cite{bbn20} & - & - & - & 69.70  \\
  RIDE~\cite{WangXD21RIDE}  & 76.52 & 74.23 & 70.45  & 72.80 \\   
  ACE~\cite{Cai2021ACE}$^\star$ & - & - & - & \textbf{72.90} \\
  RIDE+ CMO$^\star$~\cite{ParkS2022Majority} & 68.70 & 72.60 & 73.10 & 72.80 \\ 
  ResLT~\cite{Cui2022reslt}$^\star$  & 64.85 & 70.64 & 72.11 & 70.69 \\  
  \hdashline 
  RIDE+H2T & 75.69 &  74.22 & 71.36  & \underline{\textbf{73.11}} \\ 
  ResLT with H2T & 68.41 & 72.31 & 72.09 & 71.88  \\  
  \hline
 \end{tabular}
    \begin{tablenotes}
    \item \textbf{Note}: 
    The backbone is ResNet-50. Others are the same as Table~\ref{tab:cifar_results}.
    \end{tablenotes}
 \end{threeparttable}
 }
\end{table}

\begin{table}[tb]
\renewcommand{\thefootnote}{\fnsymbol{footnote}}
 \centering  
 \caption{Comparison results on Places-LT.} \label{tab:pla}  
 \resizebox{0.93\linewidth}{!}
  {\begin{threeparttable}
  \begin{tabular}{|l |c c c |c|}
  \hline 
  \multicolumn{1}{|c|}{Method}  &Head &Med &Tail & All \\
  \hline 
  \multicolumn{5}{|c|}{Single Model} \\
  \hline 
  CE loss  & 46.48  &25.66  &8.09  & 29.43 \\  
  DR~\cite{decouple20}  & 41.66 &37.79 &  32.77 & 37.40\\
  MisLAS~\cite{mislas21} & 41.95 & 41.88 & 34.65 & 40.38 \\ 
  MBJ$^\star$~\cite{Liu2022Memory}  & 39.50 & 38.20 & 35.50 & 38.10  \\   
  GCL~\cite{LiMK2022GCL}  & 38.64 &  42.59 & 38.44 &  40.30  \\
  ABL$^\star$~\cite{jin2023optimal} & 41.50 & 40.80 & 31.40 & 39.40 \\ 
  \hdashline 
  DR+H2T & 41.96 & 42.87 & 35.33 & \textbf{40.95} \\ 
  MisLAS+H2T & 41.40 & 43.04 & 35.95 & \underline{\textbf{41.03}} \\  
  GCL+H2T  & 39.34 & 42.50 & 39.46 & 40.73 \\   
  \hline
    \multicolumn{5}{|c|}{Multi-Expert Model} \\
  \hline
  LFME$^\star$~\cite{XiangLY2020LFME}  & 39.30 & 39.60 & 24.20 & 36.20 \\  
  RIDE~\cite{WangXD21RIDE} (LC) & 44.79 & 40.69 & 31.97  & 40.32 \\ 
  RIDE~\cite{WangXD21RIDE} (CC)  & 44.38 & 40.59 & 32.99 & 40.35   \\ 
  ResLT$^\star$~\cite{Cui2022reslt}  & 40.30 & 44.40 & 34.70 & 41.00 \\ 
  \hdashline 
  RIDE+H2T (LC)  & 42.99 & 42.55 & 36.25 & \underline{\textbf{41.38}} \\
  RIDE+H2T (CC) & 42.34 & 43.21 & 35.62 & \textbf{41.30} \\   
  \hline
 \end{tabular}
    \begin{tablenotes}
    \small 
    \item \textbf{Note}: 
    CC, cosine classifier. LC, linear classifier. The backbone is ResNet-152.
    Others are the same as Table~\ref{tab:cifar_results}.
    \end{tablenotes}
 \end{threeparttable}
 }
\end{table}

\noindent\textbf{Comparison Results:}\label{sec:resluts}
The results on CIFAR100-LT are shown in Table~\ref{tab:cifar_results}. 
H2T can further improve both single and multi-expert models.
Applying H2T to the vanilla training of DR with CE loss (DR+H2T) enhances the performance by a significant margin, surpassing most recent methods.
The biggest improvement exceeds 2\%.
H2T can also further boost the performance of two-stage SOTA methods, i.e., MisLAS and GCL, albeit with less significant improvements than for DR.
H2T also bolsters the efficacy of multi-expert models. For example, RIDE and ResLT integrating with H2T outperform the original methods by 1.01\% and 0.87\% on CIFAR100-LT with $\rho = 100$. 
 
The results on large-scale datasets are summarized in Tables~\ref{tab:img}, \ref{tab:iNat}, and \ref{tab:pla}. 
The baseline (CE loss) achieves high accuracy on frequent classes while the performance on rare classes is unsatisfactory. 
Comparatively, all methods, except CMO and multi-expert models, exhibit substantial improvements in the medium and tail classes but at the expense of reduced accuracy in head classes.
CMO leverages the background of head classes to augment tail classes without reducing the number of training samples in head classes, thus enhancing model performance for both head and tail classes. 
However, CMO+CE shows less competitive results in improving tail class performance.
ABL also adjusts the decision boundary, but its effectiveness is not as pronounced as H2T.
It is worth noting that all comparison methods necessitate training from scratch.
H2T can outperform the comparison methods on most datasets by re-fining the classifier with only basic CE loss and several training epochs.
For example, on iNaturalist 2018, DR+H2T achieves 71.81\%, which is the highest among all other comparison methods.
Furthermore, H2T can deliver consistent performance improvements for both single and multi-expert models, particularly for tail and medium classes.
Nevertheless, the accuracy improvements of cosine classifier (CC) are less pronounced than those of linear classifier (LC). 
For example, we can observe that H2T improves GCL using CC much less than DR and MisLAS that use LC. 
One reason is that the soft margin in GCL allowing samples to fall in the margin between classes can alleviate the decision boundary bias compared with hard margins such as LDAM~\cite{li2023adjusting}.
This phenomenon also confirms our motivation.
Table~\ref{tab:pla} further compares the improvements of H2T on RIDE with LC (by 1.06\%) and CC (0.95\%). 
The margin on LC is more profound.

\subsection{Further Analysis}\label{sec:ablation}
This section visualizes the decision boundary of H2T on each class and investigates the effects of the sampler for the fusing branch and the fusion ratio $p$ mentioned in Section~\ref{sec:H2T_method}. 
All experiments are performed with DR+H2T. 

\begin{figure}[t]
    \centering   
    \includegraphics[width=0.48\linewidth,height=0.41\linewidth]{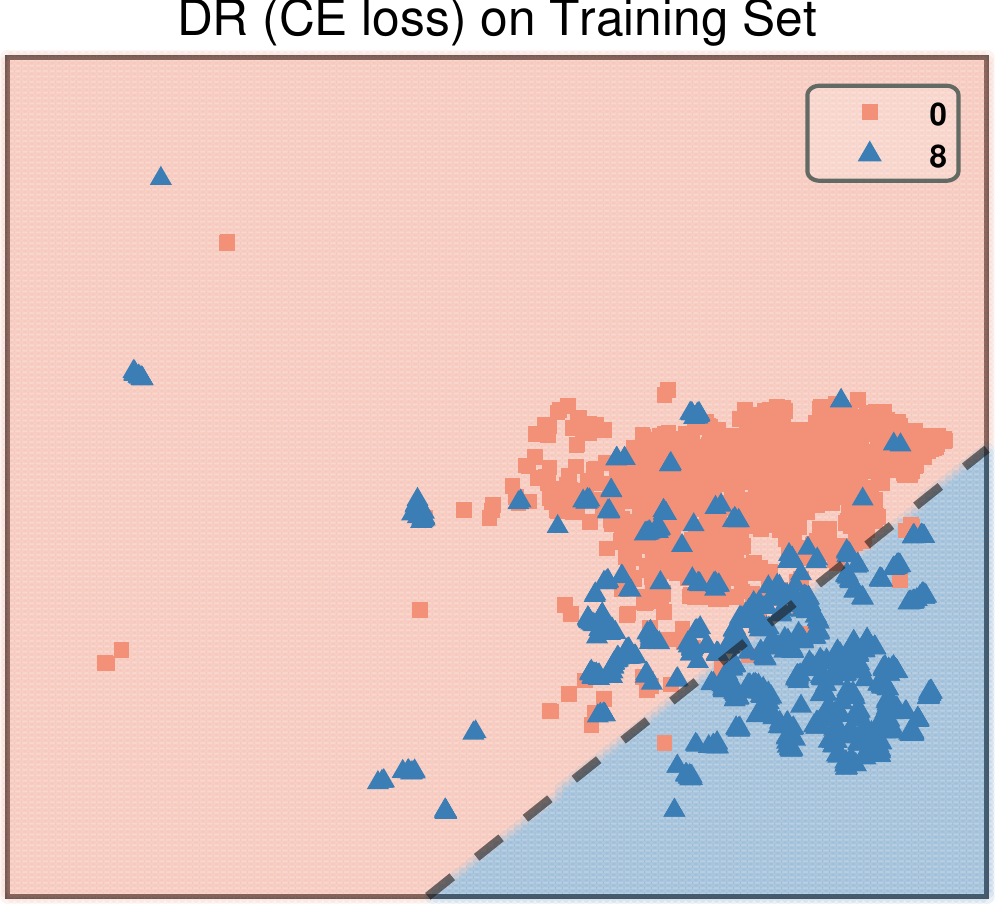} 
    \includegraphics[width=0.48\linewidth,height=0.41\linewidth]{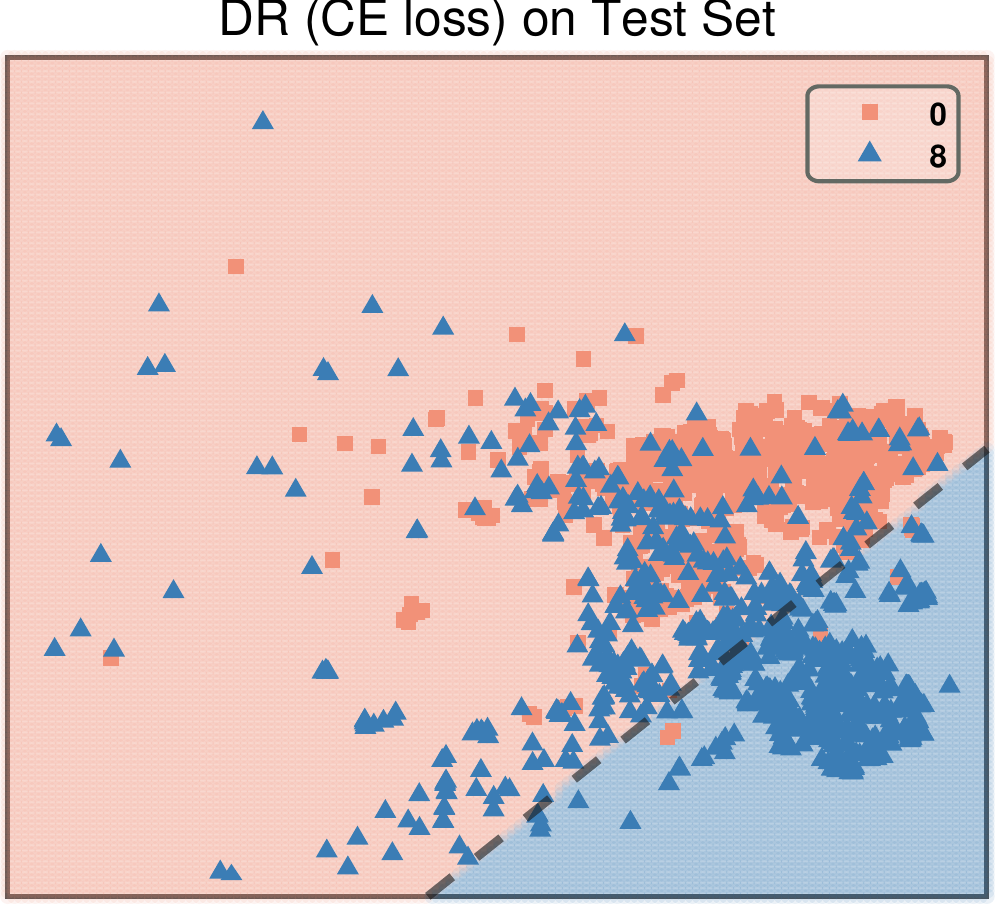} 
    \\
    \includegraphics[width=0.48\linewidth,height=0.41\linewidth]{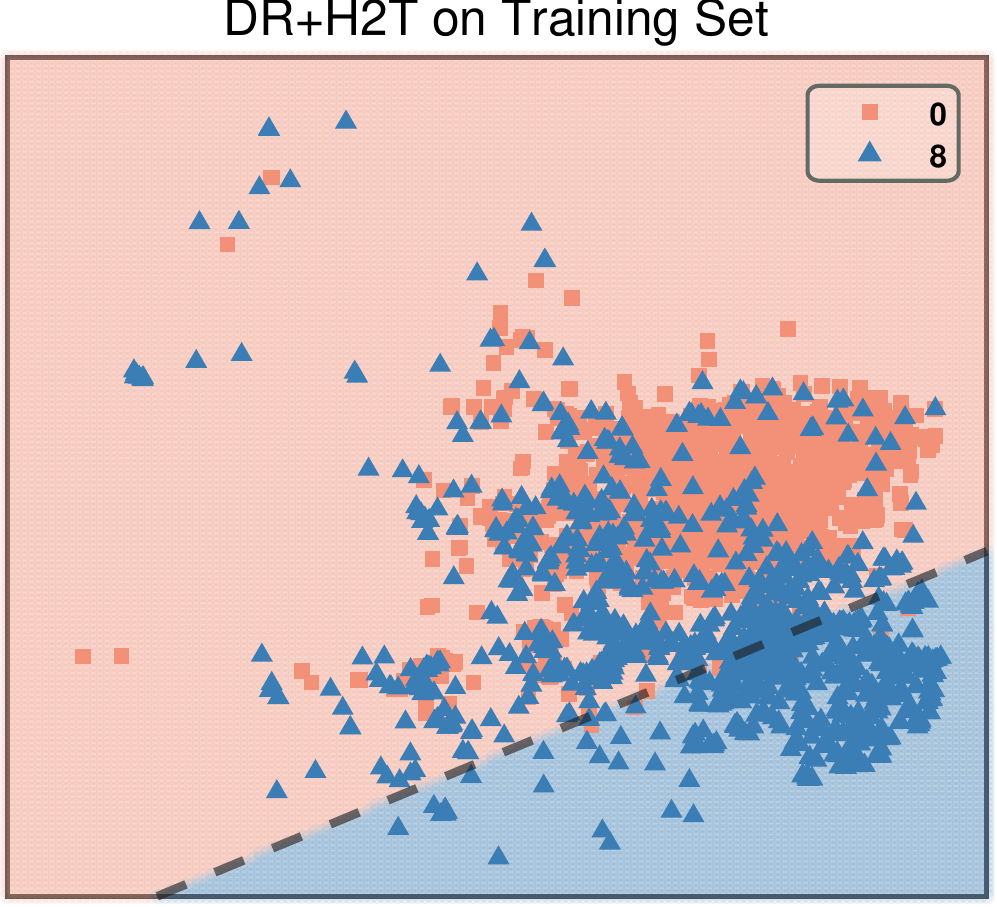}  
    \includegraphics[width=0.48\linewidth,height=0.41\linewidth]{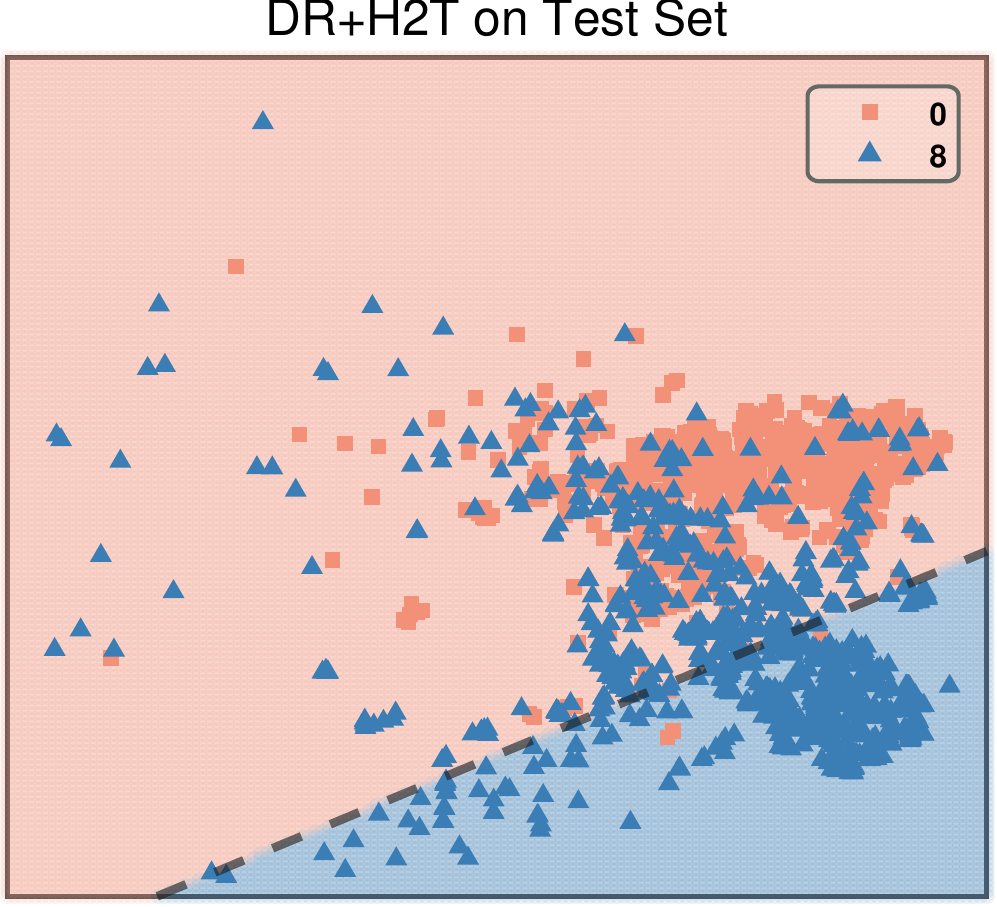}
\caption{Decision boundary comparison of Class 0 and 8 without H2T v.s. with H2T (namely, DR and DR+H2T).}
\label{fig:abl_tsne}
\end{figure}

\noindent\textbf{Visualization of Decision Boundary:}\label{sec:vis}
Figure~\ref{fig:abl_tsne} shows the t-SNE visualization of the distribution in embedding space and the decision boundary, which demonstrates our motivation (i.e., H2T can enrich tail classes and calibrate the decision boundary).
For a more convenient and clearer presentation, the experiment is conducted on CIFAR10-LT and we show the most easily misclassified classes shown in Figure~\ref{fig:dist} (i.e., Class 0 and Class 8). 
More visualization results for other classes can be found in Section~\ref{A_sec:boudary} in the Appendix.
Features are extracted from class-balanced sampling data. 
We can see that the distribution of Class 8 is sparser than that of Class 0 by DR.
H2T enriches the diversity of tail classes without external information, achieving a more optimal decision boundary.
It is worth noting that, the significance of a clear margin differs between balanced and imbalanced datasets.
In a balanced dataset, a clear margin is superior because the classifier is unbiased for each class. 
However, in the case of imbalanced training data, 
a clear margin provides the classifier more room to squeeze tail classes so that correctly classifies head classes.
Therefore, we use H2T to fill the marge with semantic samples, which has multiple advantages: 
$\left.1\right)$ prevent over-squeezing for tail classes, thus achieving a more reasonable decision boundary,
$\left.2\right)$ simulate potentially unseen samples, thus improving the generalization performance of the model on the test set.
This is distinctly different from the perception that clear margins can improve performance on balanced datasets. 

\begin{figure}[t]
    \centering
    \includegraphics[width=0.85\linewidth, height=0.56\linewidth]{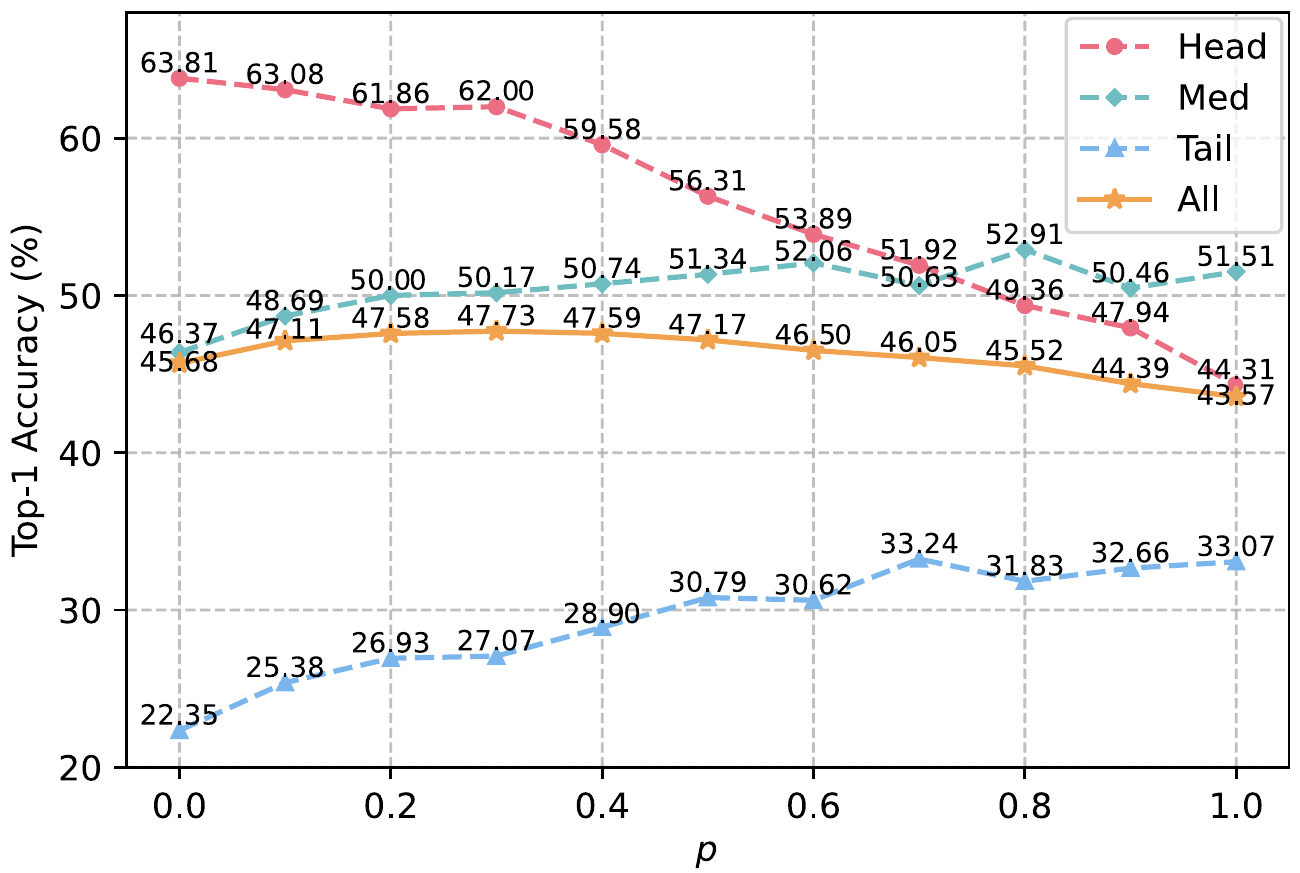}
    \caption{Change of accuracy on different splits w.r.t. $p$.} 
\label{fig:acc}
\end{figure}

\noindent\textbf{The Influence of Sampler for Fusing Branch:}
We compare different sampling strategies for the fusing branch, including class-balanced sampling (CS), reverse sampling (RS) that is tail-biased, and instance-wise sampling (IS) that is head-biased.
The results are shown in Table~\ref{tab:abla_sampler}.
DR uses cRT to fine-tune the classifier with the class-balanced sampling data, which enlarges the medium and tail class expands by pushing the decision boundary in the direction of compressing head classes.
Additionally, DR uses CE loss that does not consider class margin.
These lead to a significant boost in medium and tail classes while reducing the classification accuracy on head classes.
RS augments classes of all scales using head class samples with the lowest probability.
CS makes all classes augmented with other classes in an equal probability.
IS allows more features from the medium and tail classes to be fused with head features.
All sampling strategies adjust the decision boundaries towards narrowing the head class distribution.
RS has less damage to the performance of head classes while the improvement brought about by IS to the tail and medium classes is more prominent.
As the classifier tuning stage cannot change the embedding distribution, the performance on medium and tail classes is further improved but with performance degradation of head classes.

\begin{table}[tb]
\renewcommand{\thefootnote}{\fnsymbol{footnote}}
 \centering  
 \caption{Impact of different samplers.} 
 \label{tab:abla_sampler}
 \resizebox{0.75\linewidth}{!}
 {
  \begin{tabular}{|l |c c c |c|}
  \hline 
  \multicolumn{1}{|c|}{Method}  &Head &Med &Tail & All \\
  \hline 
  CE loss$^\dagger$ & 67.86 & 38.51 & 5.66 & 39.55 \\
  DR w.o. H2T$^\dagger$   & 63.67 & 46.63 & 22.21 & 45.68\\ 
  \hdashline 
  BS+RS & \textbf{63.56} & 48.60 & 24.07 & 46.87 \\
  BS+BS & 62.72  & 49.14  & 25.93  & 47.30 \\  
  BS+IS & 62.00 & \textbf{50.17} & \textbf{27.07} & \textbf{47.73} \\  
  \hline
 \end{tabular}
 }
\end{table}

\noindent\textbf{The Impact of Fusion Ratio:}\label{sec:ratio}
Figure~\ref{fig:acc} shows the change of accuracy on different splits with respect to $p$. 
As $p$ increases, the performance of medium and tail classes improves while that of head classes decreases, which is consistent with Section~\ref{sec:ana}.
The class extension of the medium class and the tail class is expanded by eroding that of head classes through H2T fusion. 
Since $\ddot{w}_t^T\ddot{f}_h > \ddot{w}_h^T\ddot{f}_h$ plays a disproportionate role, resulting in excessive head class samples being incorrectly labeled as medium and tail classes as $p$ increases.
If $p>0.3$, the accuracy on head classes drops dramatically.
Even replacing all feature maps with another class still yields a relatively satisfactory accuracy (43.57\%), surpassing CE loss by 4.02\%.
$p=0.3$ obtains the best overall performance (47.73\%) on CIFAR100-LT.
By adjusting $p$ based on specific scenarios, one can control the accuracy of classes with different scales, which is particularly useful when additional accuracy requirements exist for different scale classes. 
More visualization results for the fused embedding space and the selection strategy of fused features can be found in Section~\ref{A_sec:p_vis} in the Appendix. %

\section{Concluding Remarks}
This paper has proposed a simple but effective H2T for augmenting tail classes by fusing the feature maps of head class samples to tail.
By virtue of this fusion operation, our proposed H2T has two-fold advantages:
$\left.1\right)$ it produces relatively abundant features to augment tail classes, and 
$\left.2\right)$ generates two opposing forces that restrain each other, preventing excessive sacrifices in head class accuracy.
We have further designed a strategy that fuses the features of two branches with different sampling rates for easy implementation.
Extensive experiments have shown that the proposed H2T further improves upon SOTA methods.

Despite its effectiveness, H2T has an underlying assumption that representation learning has reached its optimum on the available data. 
It only adjusts the decision boundary without altering the feature distribution. 
Ultimately, improvements in the tail classes are always accompanied by sacrificing the head class performance.
Our future work will focus on obtaining a more reasonable embedding space distribution to overcome this limitation.

\section*{Acknowledgments}
This work was supported in parts by NSFC (62306181, U21B2023, 62376233), DEGP Innovation Team (2022KCX TD025), NSFC-RGC (N\_HKBU214/21), RGC (12201321, 12202622), and Shenzhen Science and Technology Program (KQTD20210811090044003, RCJC20200714114435012).

\bibliography{H2T_ref}

\begin{thebibliography}{64}
\providecommand{\natexlab}[1]{#1}

\bibitem[{Ando and Huang(2017)}]{shin2017deep}
Ando, S.; and Huang, C. 2017.
\newblock Deep Over-sampling Framework for Classifying Imbalanced Data.
\newblock In \emph{ECML/PKDD}, 770--785.

\bibitem[{Cai, Wang, and Hwang(2021)}]{Cai2021ACE}
Cai, J.; Wang, Y.; and Hwang, J.-N. 2021.
\newblock ACE: Ally Complementary Experts for Solving Long-Tailed Recognition in One-Shot.
\newblock In \emph{ICCV}, 112--121.

\bibitem[{Cao et~al.(2019)Cao, Wei, Gaidon, Arechiga, and Ma}]{Kaidi2019}
Cao, K.; Wei, C.; Gaidon, A.; Arechiga, N.; and Ma, T. 2019.
\newblock Learning imbalanced datasets with label-distribution-aware margin loss.
\newblock In \emph{NeurIPS}, 1567--1578.

\bibitem[{Chawla et~al.(2002)Chawla, Bowyer, Hall, and Kegelmeyer}]{Nitesh2002SMOTE}
Chawla, N.~V.; Bowyer, K.~W.; Hall, L.~O.; and Kegelmeyer, W.~P. 2002.
\newblock {SMOTE:} Synthetic Minority Over-sampling Technique.
\newblock \emph{Journal of artificial intelligence research}, 16: 321--357.

\bibitem[{Cheung, Li, and Zou(2021)}]{Mengke21FSG}
Cheung, Y.-M.; Li, M.; and Zou, R. 2021.
\newblock Facial Structure Guided GAN for Identity-Preserved Face Image De-Occlusion.
\newblock In \emph{ICMR}, 46–54.

\bibitem[{Chu et~al.(2020)Chu, Bian, Liu, and Ling}]{Peng2020Feature}
Chu, P.; Bian, X.; Liu, S.; and Ling, H. 2020.
\newblock Feature Space Augmentation for Long-Tailed Data.
\newblock In \emph{ECCV}, volume 12374, 694--710.

\bibitem[{Cubuk et~al.(2019)Cubuk, Zoph, Man{\'{e}}, Vasudevan, and Le}]{CubukED19AutoAugment}
Cubuk, E.~D.; Zoph, B.; Man{\'{e}}, D.; Vasudevan, V.; and Le, Q.~V. 2019.
\newblock AutoAugment: Learning Augmentation Strategies From Data.
\newblock In \emph{CVPR}, 113--123.

\bibitem[{Cubuk et~al.(2020)Cubuk, Zoph, Shlens, and Le}]{CubukED20Randaugment}
Cubuk, E.~D.; Zoph, B.; Shlens, J.; and Le, Q.~V. 2020.
\newblock Randaugment: Practical automated data augmentation with a reduced search space.
\newblock In \emph{CVPRW}, 3008--3017.

\bibitem[{Cui et~al.(2023)Cui, Liu, Tian, Zhong, and Jia}]{Cui2022reslt}
Cui, J.; Liu, S.; Tian, Z.; Zhong, Z.; and Jia, J. 2023.
\newblock ResLT: Residual Learning for Long-Tailed Recognition.
\newblock \emph{IEEE TPAMI}, 45(3): 3695--3706.

\bibitem[{Cui et~al.(2021)Cui, Zhong, Liu, Yu, and Jia}]{CuiJQ2021parametric}
Cui, J.; Zhong, Z.; Liu, S.; Yu, B.; and Jia, J. 2021.
\newblock Parametric contrastive learning.
\newblock In \emph{CVPR}, 715--724.

\bibitem[{Cui et~al.(2019)Cui, Jia, Lin, Song, and Belongie}]{cui2019class}
Cui, Y.; Jia, M.; Lin, T.-Y.; Song, Y.; and Belongie, S. 2019.
\newblock Class-balanced loss based on effective number of samples.
\newblock In \emph{CVPR}, 9268--9277.

\bibitem[{Deng et~al.(2019)Deng, Guo, Xue, and Zafeiriou}]{Deng2019ArcFace}
Deng, J.; Guo, J.; Xue, N.; and Zafeiriou, S. 2019.
\newblock ArcFace: Additive Angular Margin Loss for Deep Face Recognition.
\newblock In \emph{CVPR}, 4690--4699.

\bibitem[{He et~al.(2016)He, Zhang, Ren, and Sun}]{he2016deep}
He, K.; Zhang, X.; Ren, S.; and Sun, J. 2016.
\newblock Deep residual learning for image recognition.
\newblock In \emph{CVPR}, 770--778.

\bibitem[{Ho et~al.(2019)Ho, Liang, Chen, Stoica, and ~}]{HoD19PBA}
Ho, D.; Liang, E.; Chen, X.; Stoica, I.; and ~, P. 2019.
\newblock Population Based Augmentation: Efficient Learning of Augmentation Policy Schedules.
\newblock In \emph{ICML}, volume~97, 2731--2741.

\bibitem[{Hong et~al.(2021)Hong, Han, Choi, Seo, Kim, and Chang}]{Hong2021CVPR}
Hong, Y.; Han, S.; Choi, K.; Seo, S.; Kim, B.; and Chang, B. 2021.
\newblock Disentangling Label Distribution for Long-Tailed Visual Recognition.
\newblock In \emph{CVPR}, 6626--6636.

\bibitem[{Hu et~al.(2023)Hu, Cheung, Li, Lan, Zhang, and Liu}]{HuZK23Joint}
Hu, Z.; Cheung, Y.-m.; Li, M.; Lan, W.; Zhang, D.; and Liu, Q. 2023.
\newblock Joint Semantic Preserving Sparse Hashing for Cross-Modal Retrieval.
\newblock \emph{IEEE TCSVT}, 1--15.

\bibitem[{Hu et~al.(2019)Hu, Liu, Wang, Cheung, Wang, and Chen}]{Zhikai2019Triplet}
Hu, Z.; Liu, X.; Wang, X.; Cheung, Y.-m.; Wang, N.; and Chen, Y. 2019.
\newblock Triplet Fusion Network Hashing for Unpaired Cross-Modal Retrieval.
\newblock In \emph{ICMR}, 141–149.

\bibitem[{Huang et~al.(2016)Huang, Li, Loy, and Tang}]{Huang2016CVPR}
Huang, C.; Li, Y.; Loy, C.~C.; and Tang, X. 2016.
\newblock Learning Deep Representation for Imbalanced Classification.
\newblock In \emph{CVPR}.

\bibitem[{Huang et~al.(2020)Huang, Li, Loy, and Tang}]{HuangC2020Deep}
Huang, C.; Li, Y.; Loy, C.~C.; and Tang, X. 2020.
\newblock Deep Imbalanced Learning for Face Recognition and Attribute Prediction.
\newblock \emph{IEEE TPAMI}, 42(11): 2781--2794.

\bibitem[{Jin, Lang, and Lei(2023)}]{jin2023optimal}
Jin, L.; Lang, D.; and Lei, N. 2023.
\newblock An Optimal Transport View of Class-Imbalanced Visual Recognition.
\newblock \emph{International Journal of Computer Vision}, 1--19.

\bibitem[{Jin et~al.(2023)Jin, Li, Lu, Cheung, and Wang}]{Jin2023shike}
Jin, Y.; Li, M.; Lu, Y.; Cheung, Y.-m.; and Wang, H. 2023.
\newblock Long-Tailed Visual Recognition via Self-Heterogeneous Integration With Knowledge Excavation.
\newblock In \emph{Proceedings of the IEEE/CVF Conference on Computer Vision and Pattern Recognition (CVPR)}, 23695--23704.

\bibitem[{Kang et~al.(2020)Kang, Xie, Rohrbach, Yan, Gordo, Feng, and Kalantidis}]{decouple20}
Kang, B.; Xie, S.; Rohrbach, M.; Yan, Z.; Gordo, A.; Feng, J.; and Kalantidis, Y. 2020.
\newblock Decoupling representation and classifier for long-tailed recognition.
\newblock In \emph{ICLR}.

\bibitem[{Kim, Jeong, and Shin(2020)}]{kim2020m2m}
Kim, J.; Jeong, J.; and Shin, J. 2020.
\newblock M2m: Imbalanced classification via major-to-minor translation.
\newblock In \emph{CVPR}, 13896--13905.

\bibitem[{Krizhevsky, Hinton et~al.(2009)}]{krizhevsky2009learning}
Krizhevsky, A.; Hinton, G.; et~al. 2009.
\newblock Learning multiple layers of features from tiny images.
\newblock \emph{Tech Report}.

\bibitem[{Lan et~al.(2024)Lan, Cheung, Jiang, Hu, and Li}]{LanWC24pami}
Lan, W.; Cheung, Y.-M.; Jiang, J.; Hu, Z.; and Li, M. 2024.
\newblock Compact Neural Network via Stacking Hybrid Units.
\newblock \emph{IEEE TPAMI}, 46(1): 103--116.

\bibitem[{Li et~al.(2022{\natexlab{a}})Li, Han, Li, Fu, and Zhang}]{LiBL2022Trustworthy}
Li, B.; Han, Z.; Li, H.; Fu, H.; and Zhang, C. 2022{\natexlab{a}}.
\newblock Trustworthy Long-Tailed Classification.
\newblock In \emph{CVPR}, 6970--6979.

\bibitem[{Li et~al.(2022{\natexlab{b}})Li, Tan, Wan, Lei, and Guo}]{liJ2022nested}
Li, J.; Tan, Z.; Wan, J.; Lei, Z.; and Guo, G. 2022{\natexlab{b}}.
\newblock Nested Collaborative Learning for Long-Tailed Visual Recognition.
\newblock In \emph{CVPR}, 6949--6958.

\bibitem[{Li, Cheung, and Hu(2022)}]{LiMK2022KPS}
Li, M.; Cheung, Y.-m.; and Hu, Z. 2022.
\newblock Key Point Sensitive Loss for Long-tailed Visual Recognition.
\newblock \emph{IEEE TPAMI}, in Press: 1--14.

\bibitem[{Li, Cheung, and Lu(2022)}]{LiMK2022GCL}
Li, M.; Cheung, Y.-m.; and Lu, Y. 2022.
\newblock Long-Tailed Visual Recognition via Gaussian Clouded Logit Adjustment.
\newblock In \emph{CVPR}, 6929--6938.

\bibitem[{Li et~al.(2023)Li, Cheung, Lu, Hu, Lan, and Huang}]{li2023adjusting}
Li, M.; Cheung, Y.-m.; Lu, Y.; Hu, Z.; Lan, W.; and Huang, H. 2023.
\newblock Adjusting Logit in Gaussian Form for Long-Tailed Visual Recognition.
\newblock \emph{arXiv preprint arXiv:2305.10648}.

\bibitem[{Li et~al.(2021)Li, Gong, Liu, Wang, Qiao, and Cheng}]{Li2021MetaSAug}
Li, S.; Gong, K.; Liu, C.~H.; Wang, Y.; Qiao, F.; and Cheng, X. 2021.
\newblock {MetaSAug}: Meta Semantic Augmentation for Long-Tailed Visual Recognition.
\newblock In \emph{CVPR}, 5212--5221.

\bibitem[{Lim et~al.(2019)Lim, Kim, Kim, Kim, and Kim}]{LimS19FastAA}
Lim, S.; Kim, I.; Kim, T.; Kim, C.; and Kim, S. 2019.
\newblock Fast AutoAugment.
\newblock In \emph{NeurIPS}, 6662--6672.

\bibitem[{Lin et~al.(2020)Lin, Goyal, Girshick, He, and Doll{\'{a}}r}]{Tsung2020Focal}
Lin, T.; Goyal, P.; Girshick, R.~B.; He, K.; and Doll{\'{a}}r, P. 2020.
\newblock Focal Loss for Dense Object Detection.
\newblock \emph{IEEE TPAMI}, 42(2): 318--327.

\bibitem[{Liu, Li, and Sun(2022)}]{Liu2022Memory}
Liu, J.; Li, W.; and Sun, Y. 2022.
\newblock Memory-Based Jitter: Improving Visual Recognition on Long-Tailed Data with Diversity in Memory.
\newblock In \emph{AAAI}, 1720--1728. {AAAI} Press.

\bibitem[{Liu et~al.(2020)Liu, Sun, Han, Dou, and Li}]{Liu2020Deep}
Liu, J.; Sun, Y.; Han, C.; Dou, Z.; and Li, W. 2020.
\newblock Deep Representation Learning on Long-Tailed Data: A Learnable Embedding Augmentation Perspective.
\newblock In \emph{CVPR}.

\bibitem[{Liu et~al.(2021)Liu, Hu, Ling, and Cheung}]{Zhikai2021MTFH}
Liu, X.; Hu, Z.; Ling, H.; and Cheung, Y.-M. 2021.
\newblock MTFH: A Matrix Tri-Factorization Hashing Framework for Efficient Cross-Modal Retrieval.
\newblock \emph{IEEE TPAMI}, 43(3): 964--981.

\bibitem[{Liu et~al.(2019)Liu, Miao, Zhan, Wang, Gong, and Yu}]{LiuZW19LTOW}
Liu, Z.; Miao, Z.; Zhan, X.; Wang, J.; Gong, B.; and Yu, S.~X. 2019.
\newblock Large-Scale Long-Tailed Recognition in an Open World.
\newblock In \emph{CVPR}, 2537--2546.

\bibitem[{Minaee et~al.(2022)Minaee, Boykov, Porikli, Plaza, Kehtarnavaz, and Terzopoulos}]{Shervin2022Image}
Minaee, S.; Boykov, Y.; Porikli, F.; Plaza, A.; Kehtarnavaz, N.; and Terzopoulos, D. 2022.
\newblock Image Segmentation Using Deep Learning: A Survey.
\newblock \emph{IEEE TPAMI}, 44(7): 3523--3542.

\bibitem[{Park et~al.(2022)Park, Hong, Heo, Yun, and Choi}]{ParkS2022Majority}
Park, S.; Hong, Y.; Heo, B.; Yun, S.; and Choi, J.~Y. 2022.
\newblock The Majority Can Help The Minority: Context-rich Minority Oversampling for Long-tailed Classification.
\newblock In \emph{CVPR}, 6887--6896.

\bibitem[{Park et~al.(2021)Park, Lim, Jeon, and Choi}]{ParkS2021ICCV}
Park, S.; Lim, J.; Jeon, Y.; and Choi, J.~Y. 2021.
\newblock Influence-Balanced Loss for Imbalanced Visual Classification.
\newblock In \emph{ICCV}, 735--744.

\bibitem[{Reed(2001)}]{reed2001pareto}
Reed, W.~J. 2001.
\newblock The Pareto, Zipf and other power laws.
\newblock \emph{Economics letters}, 74(1): 15--19.

\bibitem[{Ren et~al.(2020)Ren, Yu, sheng, Ma, Zhao, Yi, and Li}]{RenJW2020Balanced}
Ren, J.; Yu, C.; sheng, s.; Ma, X.; Zhao, H.; Yi, S.; and Li, h. 2020.
\newblock Balanced Meta-Softmax for Long-Tailed Visual Recognition.
\newblock In \emph{NeurIPS}, volume~33, 4175--4186.

\bibitem[{Ren et~al.(2018)Ren, Zeng, Yang, and Urtasun}]{Mengye2018Learning}
Ren, M.; Zeng, W.; Yang, B.; and Urtasun, R. 2018.
\newblock Learning to Reweight Examples for Robust Deep Learning.
\newblock In \emph{ICML}, volume~80, 4331--4340.

\bibitem[{Russakovsky et~al.(2015)Russakovsky, Deng, Su, Krause, Satheesh, Ma, Huang, Karpathy, Khosla, Bernstein, Berg, and Fei-Fei}]{ILSVRC15}
Russakovsky, O.; Deng, J.; Su, H.; Krause, J.; Satheesh, S.; Ma, S.; Huang, Z.; Karpathy, A.; Khosla, A.; Bernstein, M.; Berg, A.~C.; and Fei-Fei, L. 2015.
\newblock ImageNet Large Scale Visual Recognition Challenge.
\newblock \emph{IJCV}, 115(3): 211--252.

\bibitem[{Szegedy et~al.(2015)Szegedy, Liu, Jia, Sermanet, Reed, Anguelov, Erhan, Vanhoucke, and Rabinovich}]{Szegedy2015Going}
Szegedy, C.; Liu, W.; Jia, Y.; Sermanet, P.; Reed, S.~E.; Anguelov, D.; Erhan, D.; Vanhoucke, V.; and Rabinovich, A. 2015.
\newblock Going deeper with convolutions.
\newblock In \emph{CVPR}, 1--9.

\bibitem[{Van~Horn et~al.(2018)Van~Horn, Mac~Aodha, Song, Cui, Sun, Shepard, Adam, Perona, and Belongie}]{Horn_2018_CVPR}
Van~Horn, G.; Mac~Aodha, O.; Song, Y.; Cui, Y.; Sun, C.; Shepard, A.; Adam, H.; Perona, P.; and Belongie, S.~J. 2018.
\newblock The INaturalist Species Classification and Detection Dataset.
\newblock In \emph{CVPR}, 8769--8778.

\bibitem[{Verma et~al.(2019)Verma, Lamb, Beckham, Najafi, Mitliagkas, Lopez-Paz, and Bengio}]{verma2019manifold}
Verma, V.; Lamb, A.; Beckham, C.; Najafi, A.; Mitliagkas, I.; Lopez-Paz, D.; and Bengio, Y. 2019.
\newblock Manifold mixup: Better representations by interpolating hidden states.
\newblock In \emph{ICML}, 6438--6447.

\bibitem[{Wang et~al.(2017)Wang, Xiang, Cheng, and Yuille}]{Wang2017NormFace}
Wang, F.; Xiang, X.; Cheng, J.; and Yuille, A.~L. 2017.
\newblock NormFace: L\({}_{\mbox{2}}\) Hypersphere Embedding for Face Verification.
\newblock In \emph{ACM MM}, 1041--1049.

\bibitem[{Wang et~al.(2018)Wang, Wang, Zhou, Ji, Gong, Zhou, Li, and Liu}]{wang2018cosface}
Wang, H.; Wang, Y.; Zhou, Z.; Ji, X.; Gong, D.; Zhou, J.; Li, Z.; and Liu, W. 2018.
\newblock Cosface: Large margin cosine loss for deep face recognition.
\newblock In \emph{CVPR}, 5265--5274.

\bibitem[{Wang et~al.(2021{\natexlab{a}})Wang, Lukasiewicz, Hu, Cai, and Zhenghua}]{Wang2021RSG}
Wang, J.; Lukasiewicz, T.; Hu, X.; Cai, J.; and Zhenghua, X. 2021{\natexlab{a}}.
\newblock {RSG:} {A} Simple but Effective Module for Learning Imbalanced Datasets.
\newblock In \emph{CVPR}, 3784--3793.

\bibitem[{Wang et~al.(2021{\natexlab{b}})Wang, Lian, Miao, Liu, and Yu}]{WangXD21RIDE}
Wang, X.; Lian, L.; Miao, Z.; Liu, Z.; and Yu, S.~X. 2021{\natexlab{b}}.
\newblock Long-tailed Recognition by Routing Diverse Distribution-Aware Experts.
\newblock In \emph{ICLR}.

\bibitem[{Xiang, Ding, and Han(2020)}]{XiangLY2020LFME}
Xiang, L.; Ding, G.; and Han, J. 2020.
\newblock Learning From Multiple Experts: Self-paced Knowledge Distillation for Long-Tailed Classification.
\newblock In \emph{ECCV}, 247--263.

\bibitem[{Xiao et~al.(2021)Xiao, Zhang, Jing, Huang, and Song}]{xiao2021does}
Xiao, L.; Zhang, X.; Jing, L.; Huang, C.; and Song, M. 2021.
\newblock Does head label help for long-tailed multi-label text classification.
\newblock \emph{AAAI}, 35(16): 14103--14111.

\bibitem[{Yang et~al.(2022)Yang, Jiang, Song, and Guo}]{yang2022survey}
Yang, L.; Jiang, H.; Song, Q.; and Guo, J. 2022.
\newblock A survey on long-tailed visual recognition.
\newblock \emph{IJCV}, 130(7): 1837--1872.

\bibitem[{Yin et~al.(2019)Yin, Yu, Sohn, Liu, and Chandraker}]{yin2019feature}
Yin, X.; Yu, X.; Sohn, K.; Liu, X.; and Chandraker, M. 2019.
\newblock Feature transfer learning for face recognition with under-represented data.
\newblock In \emph{CVPR}, 5704--5713.

\bibitem[{Yun et~al.(2019)Yun, Han, Oh, Chun, Choe, and Yoo}]{YunSD2019cutmix}
Yun, S.; Han, D.; Oh, S.~J.; Chun, S.; Choe, J.; and Yoo, Y. 2019.
\newblock Cutmix: Regularization strategy to train strong classifiers with localizable features.
\newblock In \emph{ICCV}, 6023--6032.

\bibitem[{Zada, Benou, and Irani(2022)}]{Zada22noise}
Zada, S.; Benou, I.; and Irani, M. 2022.
\newblock Pure Noise to the Rescue of Insufficient Data: Improving Imbalanced Classification by Training on Random Noise Images.
\newblock In \emph{ICML}, volume 162, 25817--25833.

\bibitem[{Zhang et~al.(2018)Zhang, Cisse, Dauphin, and Lopez-Paz}]{Hongyi2018}
Zhang, H.; Cisse, M.; Dauphin, Y.~N.; and Lopez-Paz, D. 2018.
\newblock mixup: Beyond empirical risk minimization.
\newblock \emph{ICLR}.

\bibitem[{Zhang et~al.(2023)Zhang, Kang, Hooi, Yan, and Feng}]{zhang2021survey}
Zhang, Y.; Kang, B.; Hooi, B.; Yan, S.; and Feng, J. 2023.
\newblock Deep Long-Tailed Learning: A Survey.
\newblock \emph{IEEE TPAMI}, 1--20.

\bibitem[{Zhang et~al.(2021)Zhang, Wei, Zhou, and Wu}]{zhang2021bag}
Zhang, Y.; Wei, X.-S.; Zhou, B.; and Wu, J. 2021.
\newblock Bag of Tricks for Long-Tailed Visual Recognition with Deep Convolutional Neural Networks.
\newblock In \emph{AAAI}, 3447--3455.

\bibitem[{Zhong et~al.(2021)Zhong, Cui, Liu, and Jia}]{mislas21}
Zhong, Z.; Cui, J.; Liu, S.; and Jia, J. 2021.
\newblock Improving Calibration for Long-Tailed Recognition.
\newblock In \emph{CVPR}, 16489--16498.

\bibitem[{Zhou et~al.(2020)Zhou, Cui, Wei, and Chen}]{bbn20}
Zhou, B.; Cui, Q.; Wei, X.-S.; and Chen, Z.-M. 2020.
\newblock {BBN}: Bilateral-Branch Network with Cumulative Learning for Long-Tailed Visual Recognition.
\newblock In \emph{CVPR}, 9719--9728.

\bibitem[{Zhou et~al.(2016)Zhou, Khosla, Lapedriza, Oliva, and Torralba}]{Zhou2016Learning}
Zhou, B.; Khosla, A.; Lapedriza, {\`{A}}.; Oliva, A.; and Torralba, A. 2016.
\newblock Learning Deep Features for Discriminative Localization.
\newblock In \emph{CVPR}, 2921--2929.

\bibitem[{Zhou et~al.(2017)Zhou, Lapedriza, Khosla, Oliva, and Torralba}]{zhou2017places}
Zhou, B.; Lapedriza, A.; Khosla, A.; Oliva, A.; and Torralba, A. 2017.
\newblock Places: A 10 million image database for scene recognition.
\newblock \emph{IEEE TPAMI}, 40(6): 1452--1464.

\end{thebibliography}

\clearpage
\twocolumn[
\begin{@twocolumnfalse}
\section*{\huge Appendix: Feature Fusion from Head to Tail for Long-Tailed Visual Recognition \\[25pt] }
\end{@twocolumnfalse}
]

\gdef\thesection{Appendix~\Alph{section}}  
\appendix
\section{Key Execution Code}

H2T is easy to implement and can be seamlessly integrated with various other methods. 
Below, we outline the key codes of the algorithm:
\begin{listing}[htpb]%
\caption{PyTorch implementation of {\tt H2T}}%
\label{lst:listing}%
\begin{lstlisting}[language=python]
def H2T(x1, x2, rho = 0.3):          
    if type(x1).__name__ == 'Tensor':
        fea_num = x1.shape[1]
        index = torch.randperm(fea_num).cuda()
        slt_num = int(rho*fea_num)
        index = index[:slt_num]
        x1[:,index,:,:] = x2[:,index,:,:] 
    else:
        for i in range(len(x1)):
            fea_num = x1[i].shape[1]
            index = torch.randperm(fea_num).cuda()
            slt_num = int(rho*fea_num)
            index = index[:slt_num]
            x1[i][:,index,:,:] = x2[i][:,index,:,:]    
    return x1
\end{lstlisting}
\end{listing}

\section{Comparison Results on CIFAR-10-LT} \label{sec:supp_cifar10}

In Section~\ref{sec:resluts} of the main paper, we show the results on most of the benchmark datasets except CIRAT10-LT.  
Table~\ref{tab:cifar10_results} presents the supplementary results on CIFAR10-LT w.r.t Top-1 classification accuracy (\%). The improvement of applying H2T to the vanilla training of DR with CE loss (DR+H2T) is more clear.
The biggest improvement exceeds 5\% on CIFAR10-LT with $\rho = 200$. 
The improvement in GCL+H2T is less compared with other methods because GCL uses the distance in cosine form. 
This result is consistent with the experimental results on Places-LT (refer to Table~\ref{tab:pla} of the main paper), namely the H2T improves RIDE with cosine classifier less than with linear classifier.
\begin{table}[tb]
\renewcommand{\thefootnote}{\fnsymbol{footnote}}
 \centering  
 \caption{Comparison results on CIFAR10-LT.} %
 \label{tab:cifar10_results}
 \resizebox{0.95\linewidth}{!}
  {\begin{threeparttable}
  \begin{tabular}{|l | c c c|}
  \hline 
  \multicolumn{1}{|c|}{Imbalance Ratio} & 200 & 100 &50 \\
  \hline 
  \multicolumn{4}{|c|}{Single Model} \\
  \hline 
  CE loss  & 67.24  &70.70  &74.81  \\
  LDAM-DRW \cite{Kaidi2019}  & 73.52  &77.03  &81.03\\
  DR+MU~\cite{decouple20}  & 73.06 &79.15 & 84.21 \\
  MisLAS~\cite{mislas21} &77.17& 82.12 & 85.55 \\
  MBJ$^\star$~\cite{Liu2022Memory}  & - & 81.00 & \textbf{86.60}  \\
  GCL~\cite{LiMK2022GCL} & 78.56 & 82.06 & 85.36 \\
  \hdashline 
  {DR+H2T} & \makecell{\textbf{78.59} \\ \color{deepblue}($\uparrow $ 5.53)} & \makecell{82.87 \\ \color{deepblue}($\uparrow $ 2.14)} & \makecell{86.35 \\ \color{deepblue}($\uparrow $ 3.72)} \\
  {MisLAS+H2T}  & \makecell{ 78.05 \\ \color{deepblue}($\uparrow $ 0.88)}  & \makecell{\underline{\textbf{83.23}} \\ \color{deepblue}($\uparrow $ 1.11)} & \makecell{\underline{\textbf{86.72}}\\ \color{deepblue}($\uparrow $ 1.17)} \\
  {GCL+H2T} & \makecell{\underline{\textbf{79.42}} \\ \color{deepblue}($\uparrow $ 0.86)} & \makecell{\textbf{82.40} \\ \color{deepblue}($\uparrow $ 0.34)} & \makecell{85.41 \\ \color{deepblue}($\uparrow $ 0.05)} \\ 
  \hline
    \multicolumn{4}{|c|}{Multi-Expert Model} \\
  \hline
  BBN~\cite{bbn20} & 73.47 & 79.82 & 81.18 \\
  ACE$^\star$~\cite{Cai2021ACE}  & - & \textbf{81.20} & \textbf{84.30} \\
  ResLT~\cite{Cui2022reslt}  & \textbf{75.29} & 80.58 & 84.07\\
  \hdashline 
  ResLT+H2T  & \makecell{\underline{\textbf{77.25}} \\ \color{deepblue}($\uparrow $ 1.96) }& \makecell{\underline{\textbf{81.77}} \\ \color{deepblue}($\uparrow $ 1.19)} & \makecell{\underline{\textbf{84.99}} \\ \color{deepblue}($\uparrow $ 0.92)}\\
  \hline
 \end{tabular}
    \begin{tablenotes}
    \item \textbf{Note}: 
    The backbone is ResNet-32.  
    $\star$ represents the results quoted from the corresponding papers.
    The best and the second-best results are shown in \underline{\textbf{underline bold}} and \textbf{bold}, respectively. 
    The {\color{deepblue}text in blue} indicates the performance improvement compared to the original method w.o. H2T.
    \end{tablenotes}
 \end{threeparttable}
 }
\end{table}

\section{Selection of fused features} \label{sec:supp_rou}

\begin{table}[tb]
 \centering  
 \caption{Selection of fused features} \label{tab:abla_slt}
 \resizebox{0.95\linewidth}{!}{
  \begin{tabular}{|c |c |c |c |}
  \hline 
  First 30\% & Middle 30\% & Last 30\% & Random 30$\%$\_1 \\
  \hline 
  47.22 & 47.23 & 47.41 & 47.73   \\
  \hline 
  Random 30\%\_2 & Random \%30\_3 & Random 30\%\_4 & Random 30\%\_5 \\
  \hline   
  47.75 & 47.80 & 47.71 & 47.66 \\
  \hline
 \end{tabular}  }
 \vspace{-12pt}
\end{table}

The selected fused features are essential. 
We conduct an experiment to show the effect of the selection strategy of fused features.
The dataset is CIRAR100-LT and the imbalance ratio is 100.
Table~\ref{tab:abla_slt} shows the Top-1 classification accuracy (\%) of different selections.
Two approaches are validated: sequential selection and random selection.
Sequential selection refers to replacing a certain proportion of feature maps directly in accordance with the original order of feature maps.
We select the first 30\%, meddle 30\%, and the last 30\% of the feature maps to be replaced.
Random selection refers to randomly selecting a certain proportion of feature maps to replace.
In the experiment, the fused ratio is 30\% and we verified 5 random seeds (Random 30\%\_1 to Random 30\%\_5).
We can observe that random selection generally outperforms sequential selection. 
Different fused features may introduce variations, underscoring the need for a thoughtful selection strategy. 
This aspect remains a topic for our future work.

\begin{figure*}

\begin{minipage}[c]{.48\linewidth}
    \centering  
    \includegraphics[width=.49\linewidth, height=.49\linewidth]{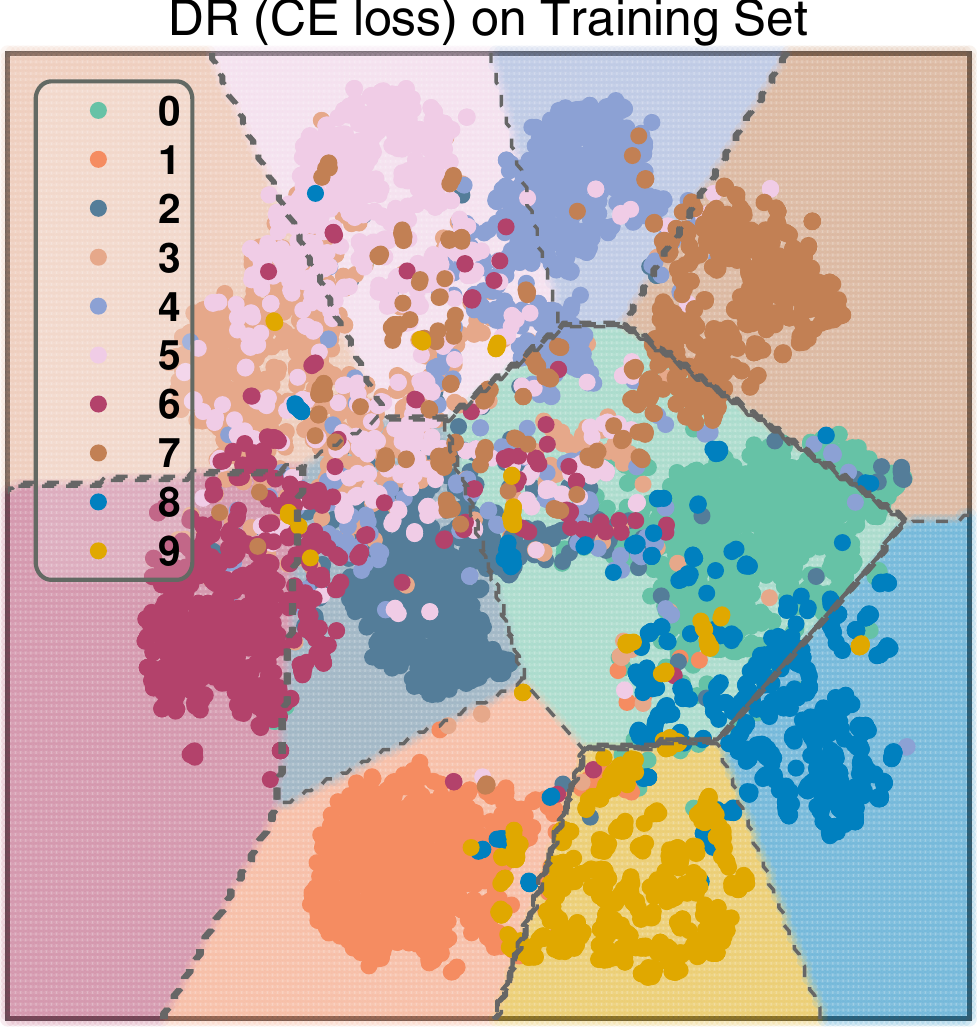}\label{fig:CE-train} 
    \includegraphics[width=.49\linewidth, height=.493\linewidth]{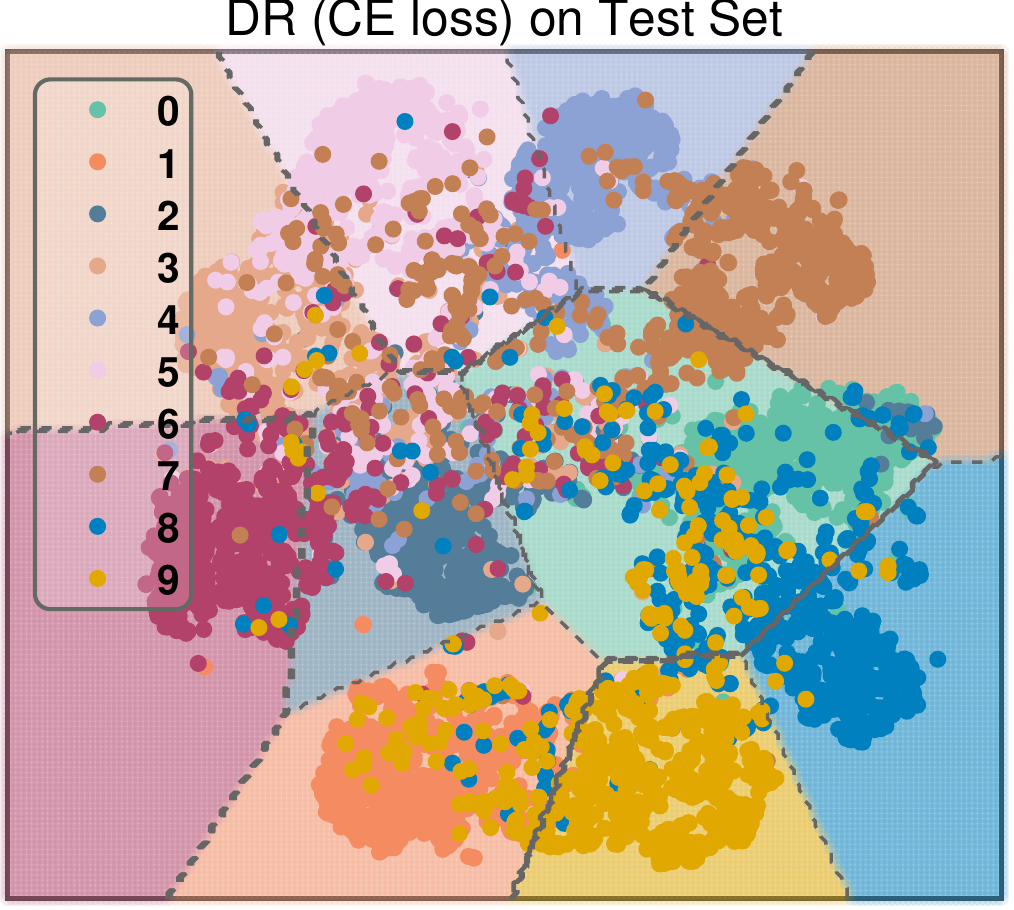}\label{fig:h2t-train}  
    \\
    \includegraphics[width=.49\linewidth, height=.49\linewidth]{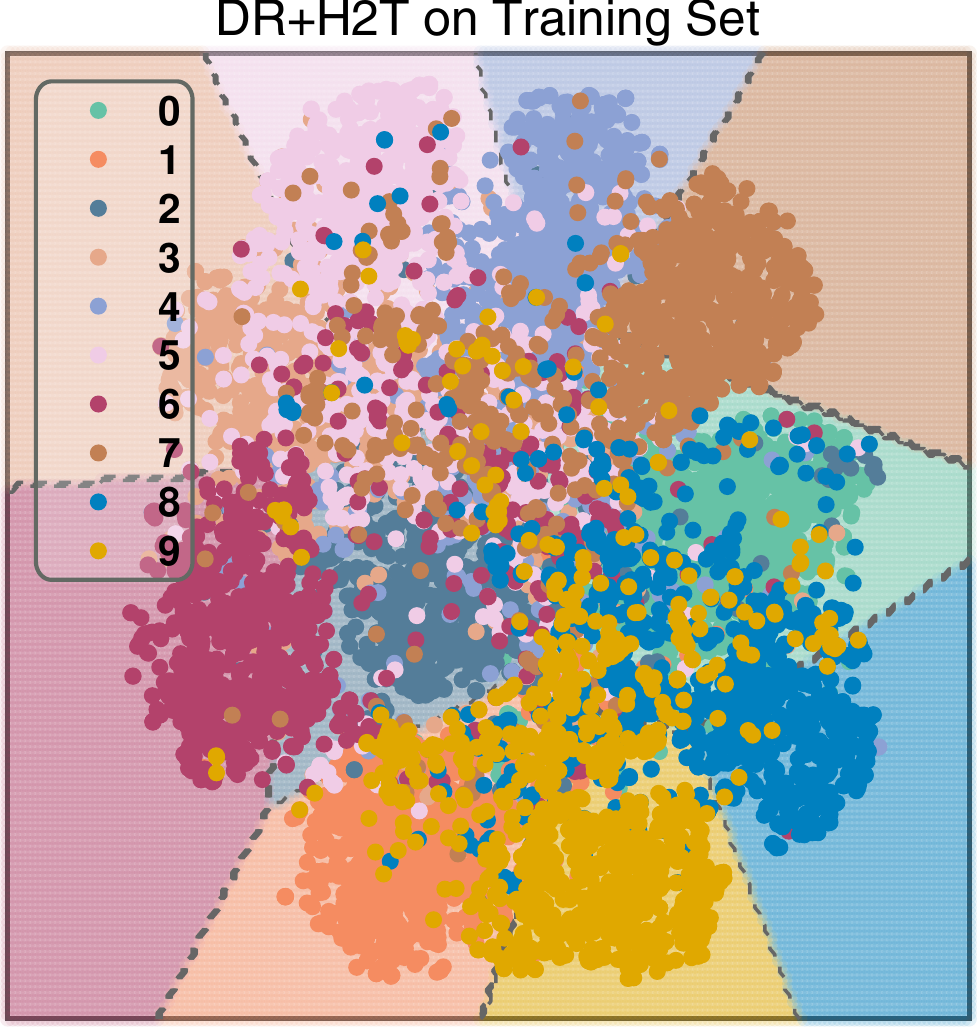}\label{fig:CE-val}  
    \includegraphics[width=.49\linewidth, height=.493\linewidth]{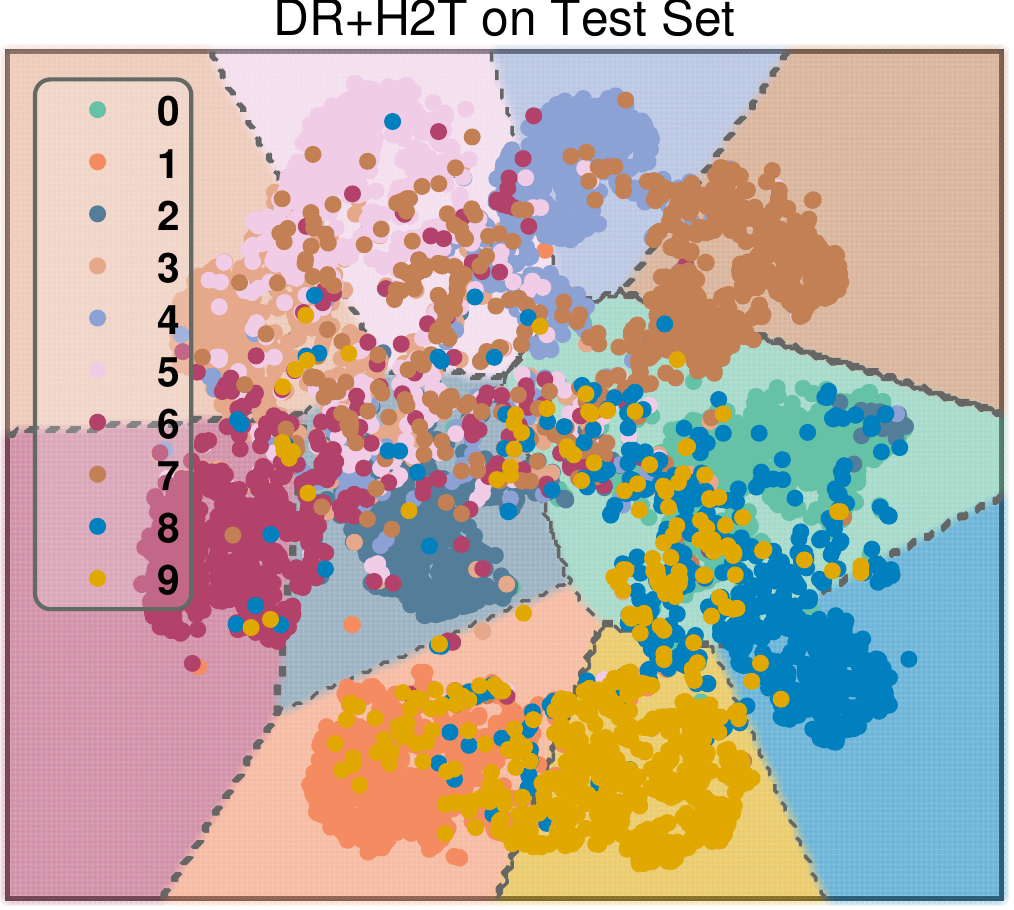}\label{fig:h2t-val}
\caption{Decision boundary Comparison w.o. and w. H2T (all classes).}
\label{fig:supp_tsne}
\end{minipage}
\hfill
\begin{minipage}[c]{.48\linewidth} 
    \centering  
    \includegraphics[width=.49\linewidth, height=.49\linewidth]{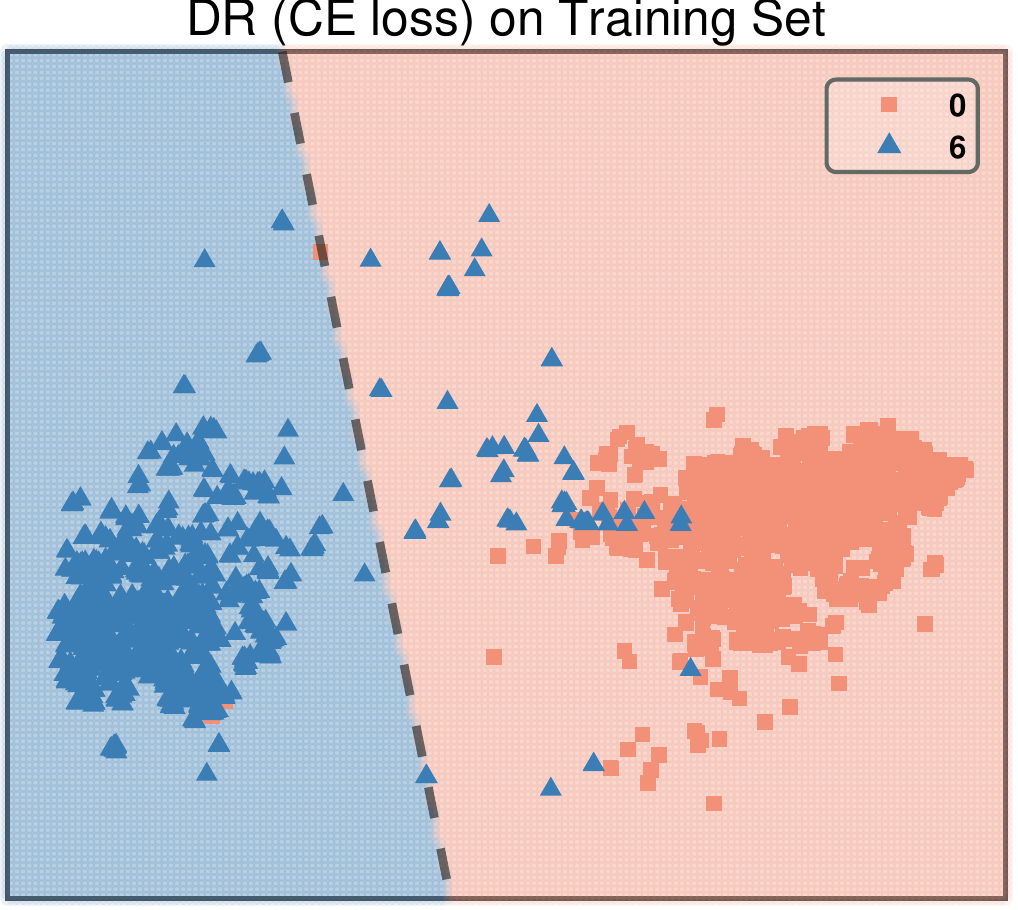}\label{fig:CE-train06}
    \includegraphics[width=.49\linewidth, height=.49\linewidth]{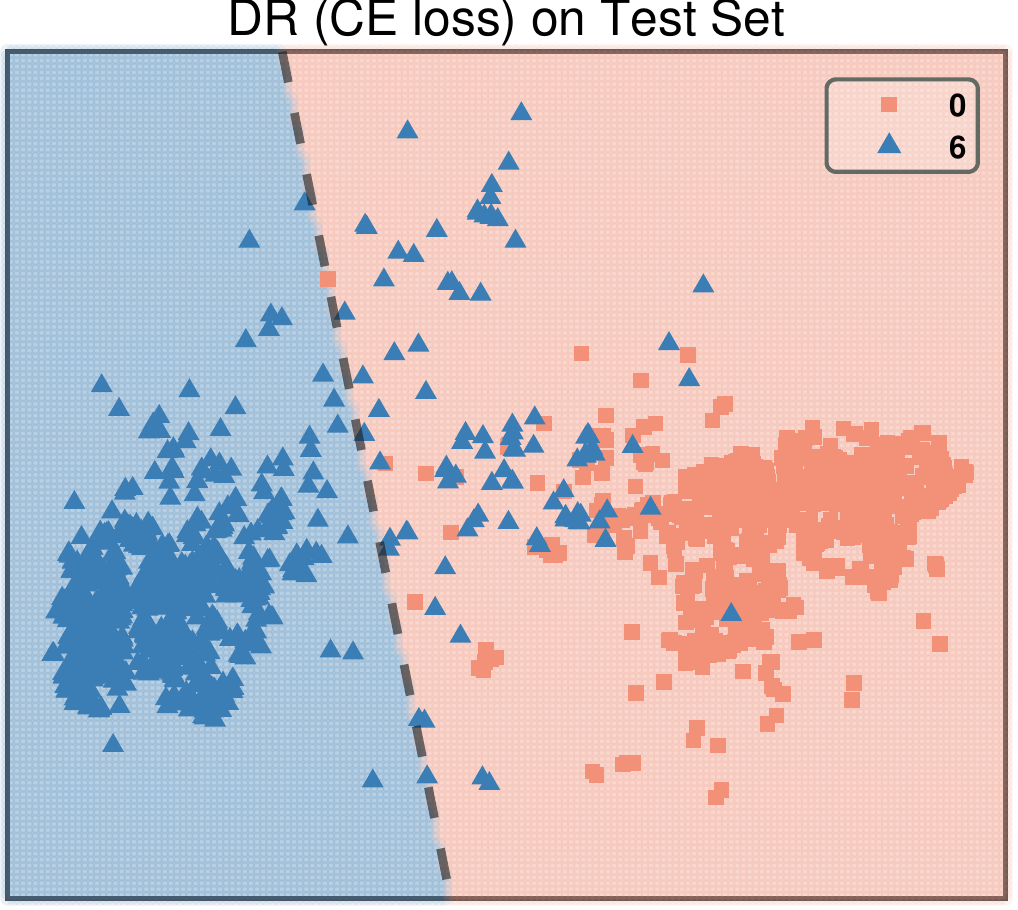}\label{fig:CE-val06}
    \\
    \includegraphics[width=.49\linewidth, height=.49\linewidth]{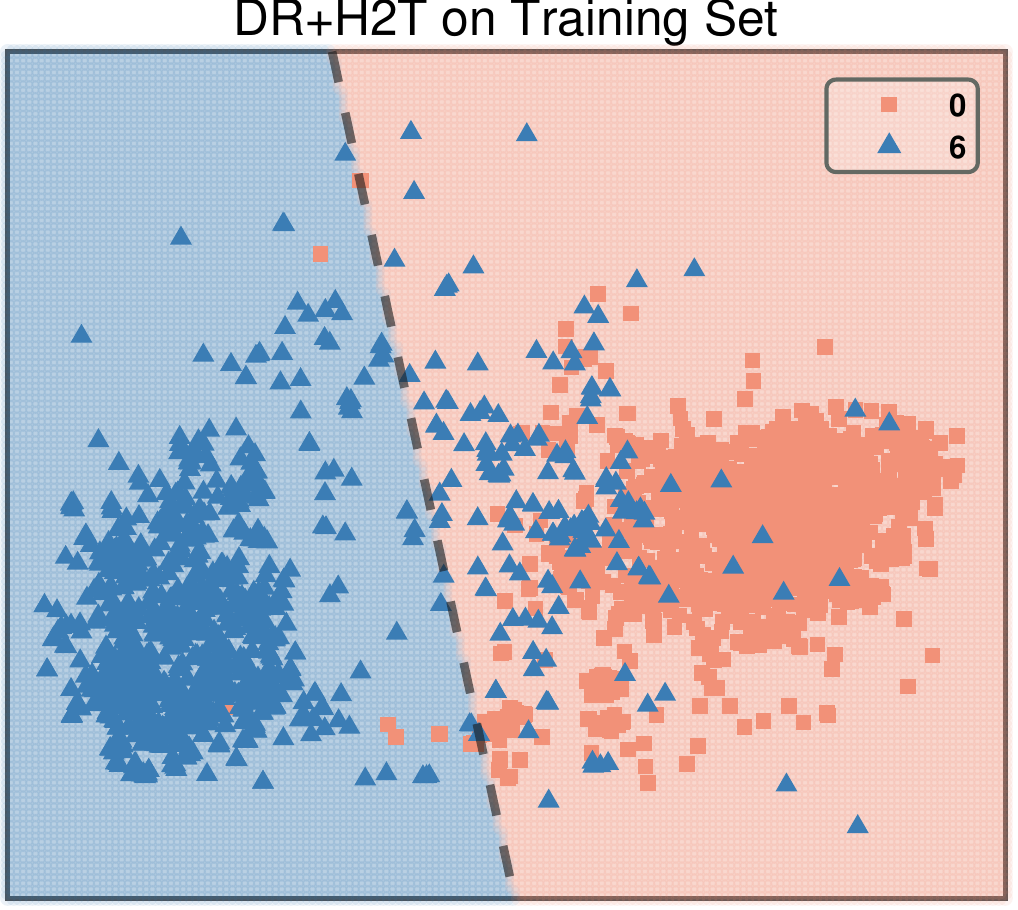}\label{fig:h2t-train06}
    \includegraphics[width=.49\linewidth, height=.49\linewidth]{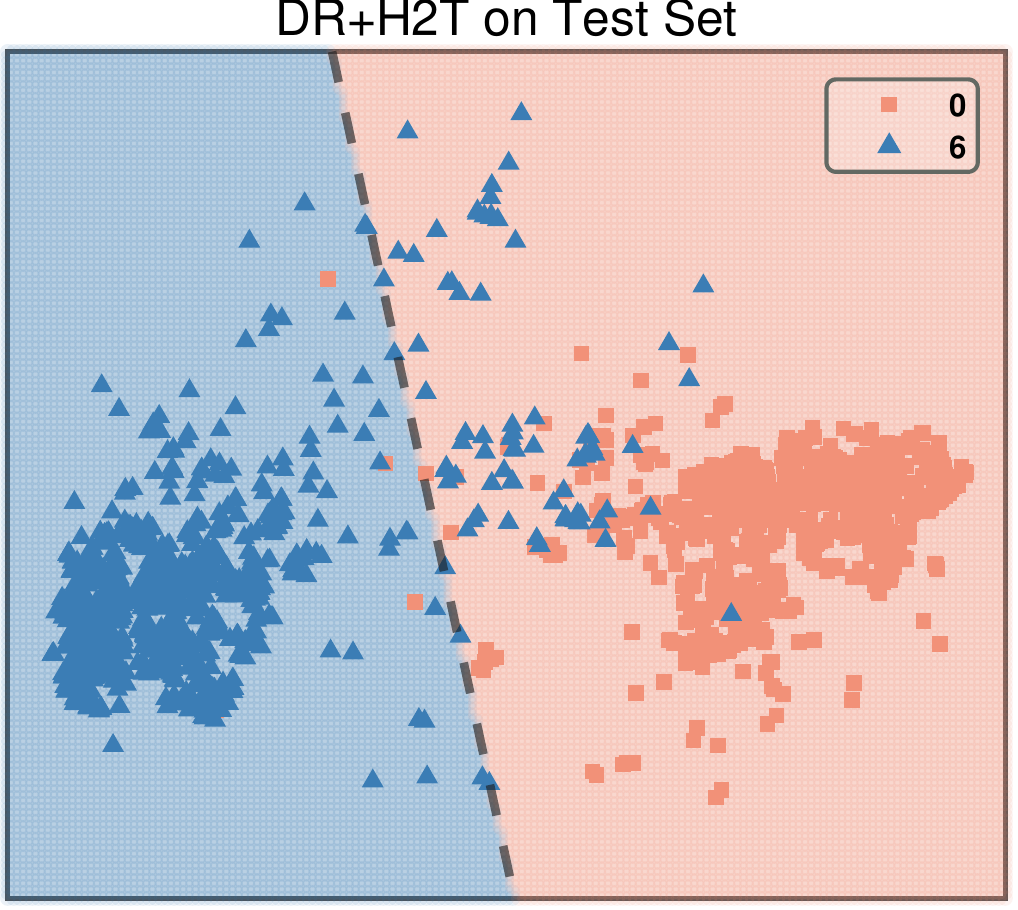}\label{fig:h2t-val06}
\caption{Decision boundary Comparison w.o. and w. H2T (Class 0 and 6).}
\label{fig:supp_tsne06}
\end{minipage}

\medskip

\begin{minipage}[c]{.48\linewidth}  
    \centering  
    \includegraphics[width=.49\linewidth, height=.49\linewidth ]{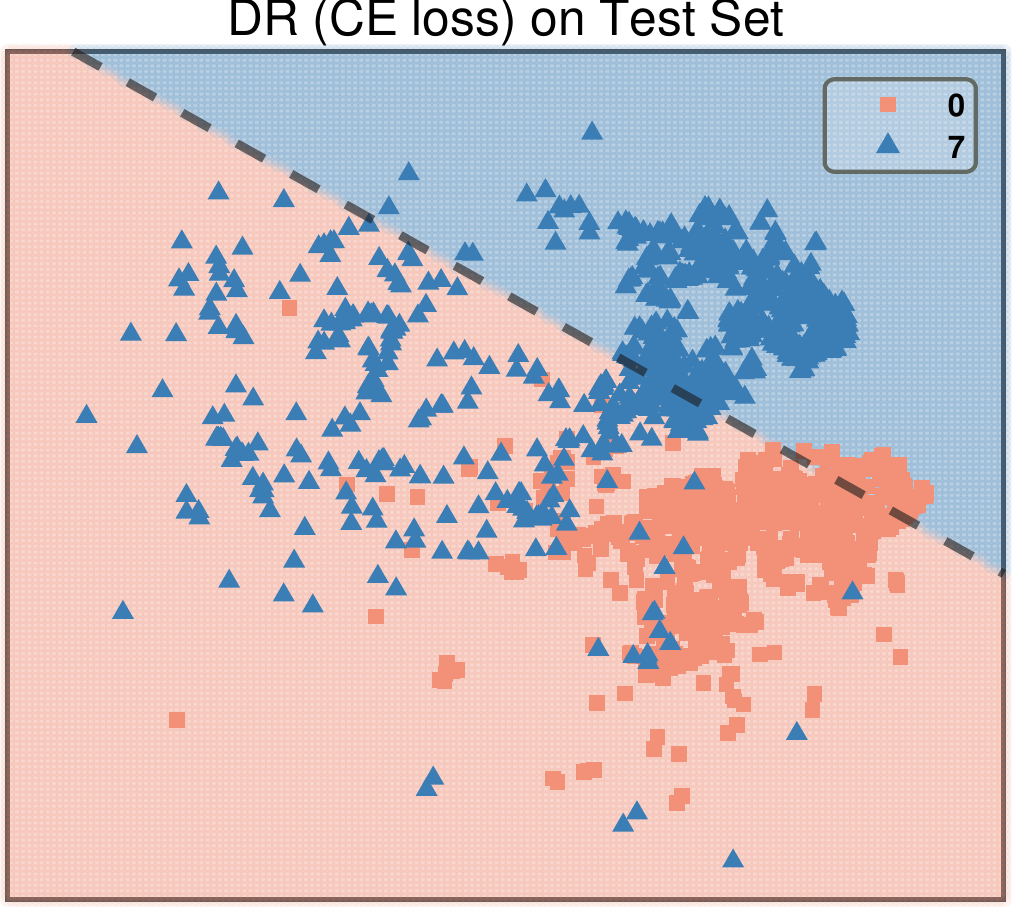}\label{fig:CE-train07} 
    \includegraphics[width=.49\linewidth, height=.49\linewidth]{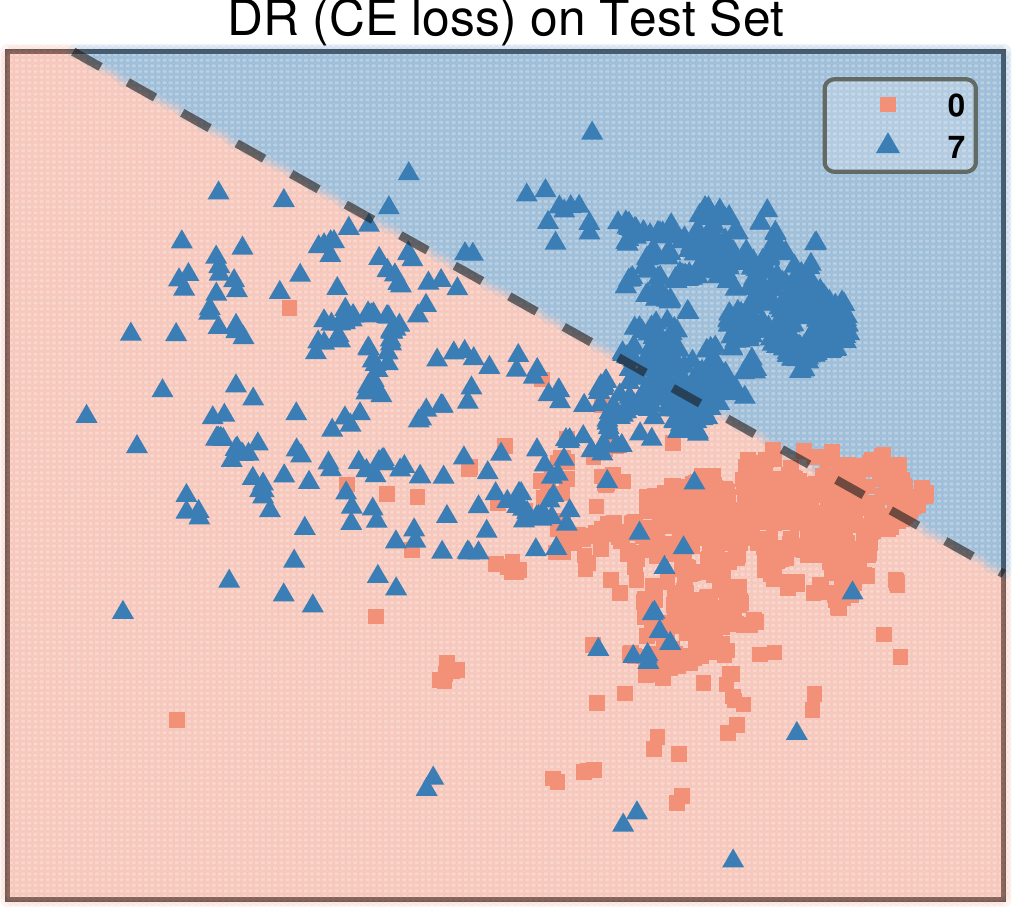}\label{fig:h2t-train07} 
    \\
    \includegraphics[width=.49\linewidth, height=.49\linewidth]{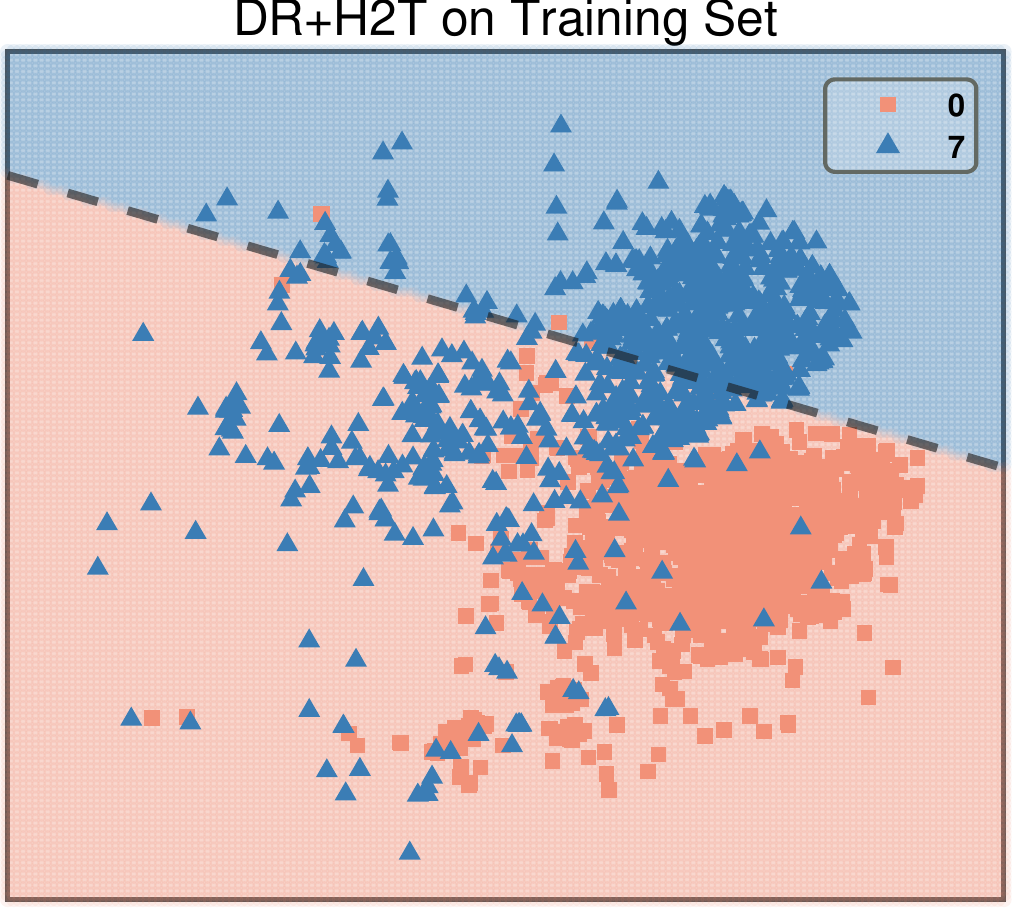}\label{fig:h2t-val07}
    \includegraphics[width=.49\linewidth, height=.49\linewidth]{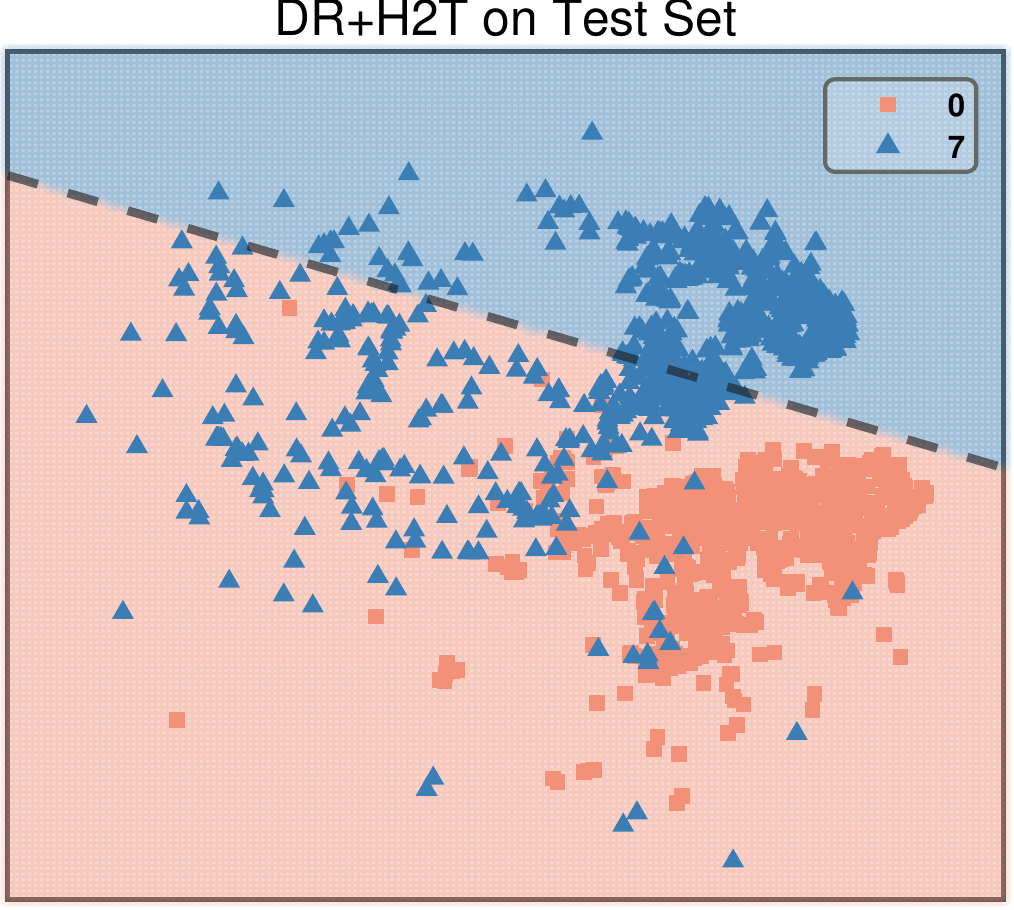}\label{fig:CE-val07} 
\caption{Decision boundary Comparison w.o. and w. H2T (Class 0 and 7).}
\label{fig:supp_tsne07}
\end{minipage}
\hfill
\begin{minipage}[c]{.48\linewidth}  
    \centering  
    \includegraphics[width=.49\linewidth, height=.49\linewidth]{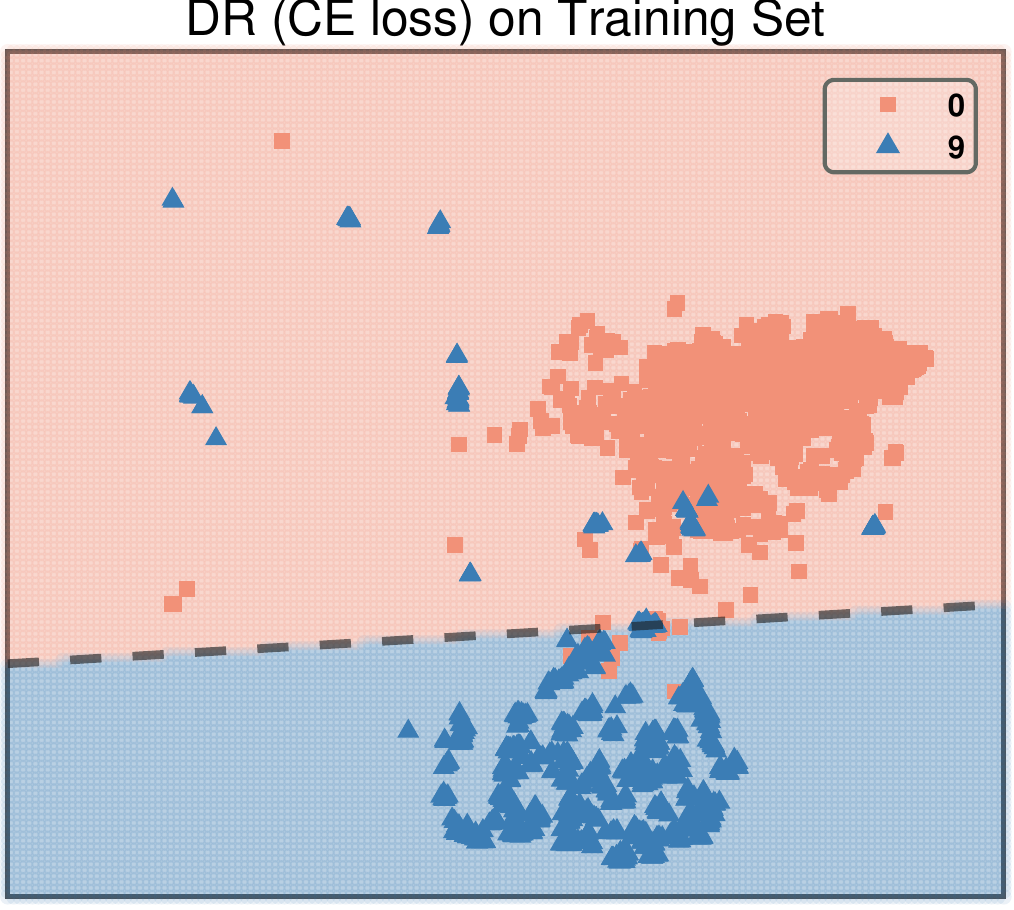}\label{fig:CE-train09} 
    \includegraphics[width=.49\linewidth, height=.49\linewidth]{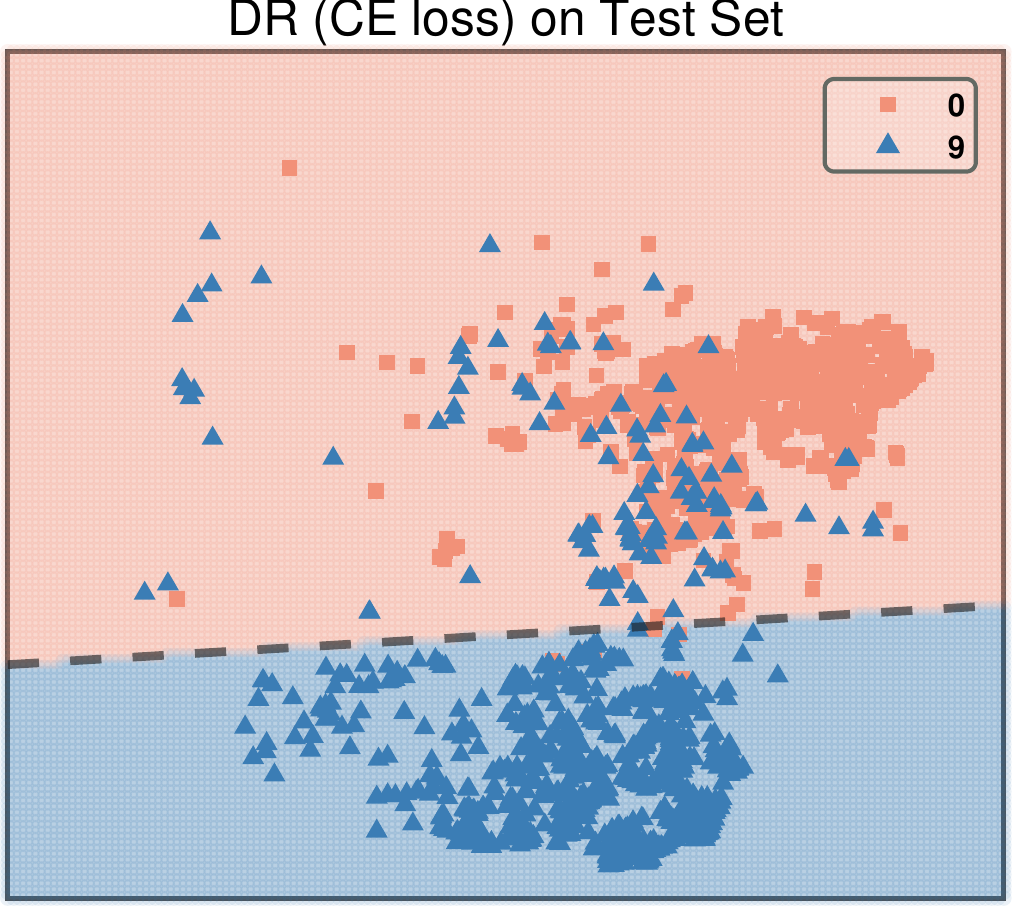}\label{fig:CE-val09} 
    \\
    \includegraphics[width=.49\linewidth, height=.49\linewidth]{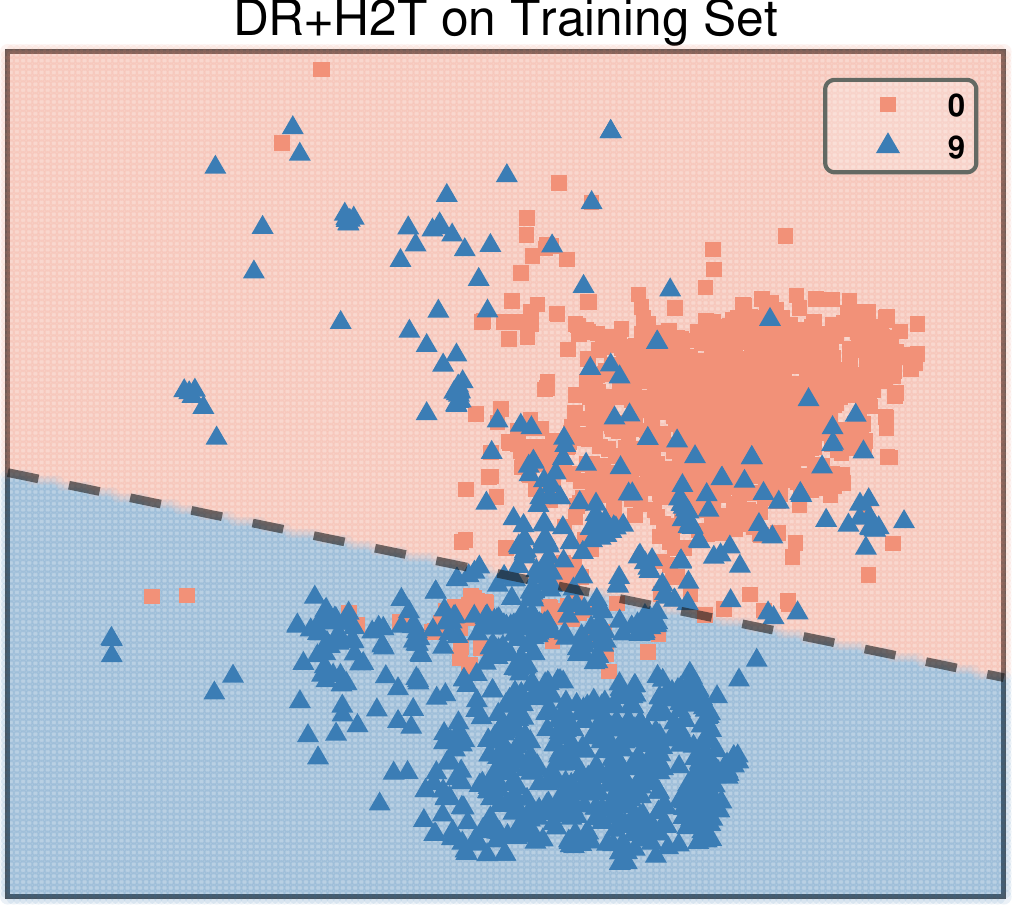}\label{fig:h2t-train09} 
    \includegraphics[width=.49\linewidth, height=.49\linewidth]{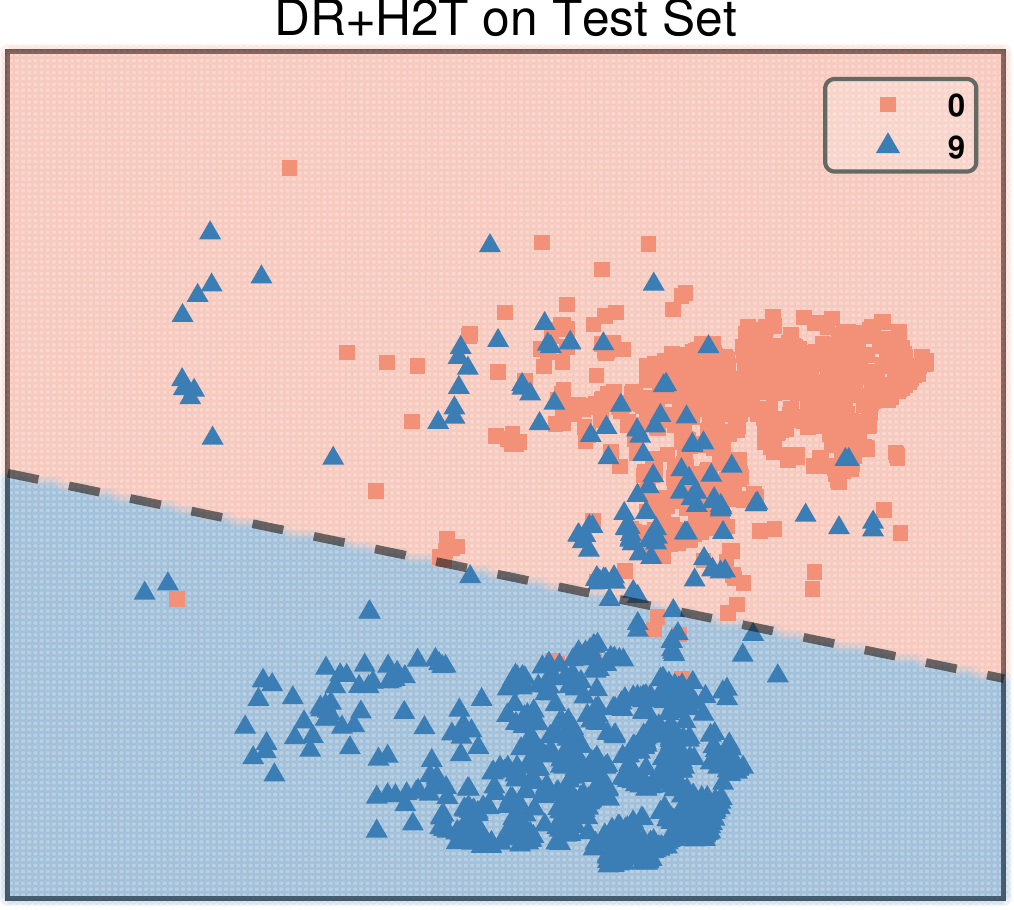}\label{fig:h2t-val09}
\caption{Decision boundary Comparison w.o. and w. H2T (Class 0 and 9).}
\label{fig:supp_tsne09}
\end{minipage}
\end{figure*}

\section{Visualization Results of decision Boundaries} \label{A_sec:boudary}

This section offers additional visualization comparisons pertaining to the decision boundary, serving as a supplementary analysis to Figure~\ref{fig:abl_tsne} presented in Section~\ref{sec:ablation} of the main paper. 
CIFAR10-LT, with an imbalance ratio of 100, is employed for the purpose of facilitating easy visualization.
Classes 0-3 are head classes and Classes 6-9 are tail classes.
Figure~\ref{fig:supp_tsne} shows the tSNE of embedding space and the decision boundaries. 
We can see that H2T changes the embedding space of head and tail classes more.
Figures~\ref{fig:supp_tsne06}-\ref{fig:supp_tsne09} further visualize the decision boundary between head and each tail class.
Compared with DR, H2T fills the margin between classes and adjusts the decision boundary to a more reasonable position in the obtained embedding space.
For example, as shown in \ref{fig:supp_tsne07}, H2T not only expands the tail class (Class 7) space but also rectifies the head class space (Class 0).
While this decision boundary enhances the classification effect, it remains suboptimal. 
This presents an intriguing problem worth exploring in future research.

\section{Visualization Results w.r.t. $p$} \label{A_sec:p_vis}
\begin{figure*}[htpb]
	\centering
        \includegraphics[width=0.31\linewidth]{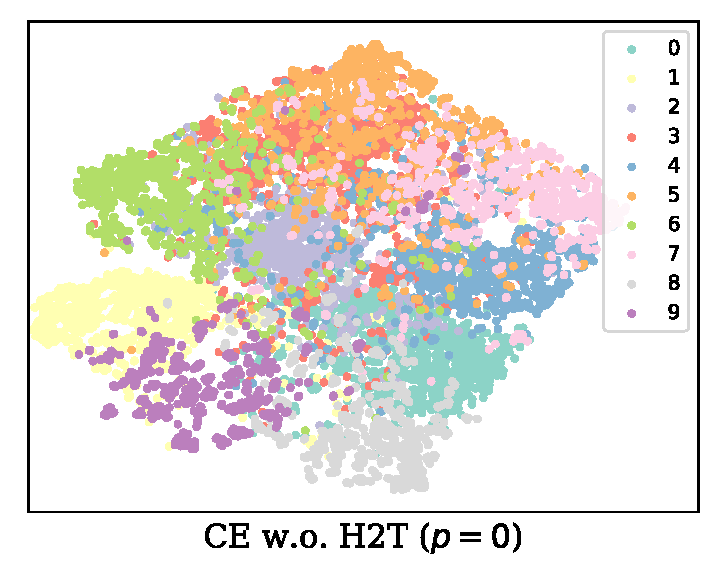}
        \includegraphics[width=0.31\linewidth]{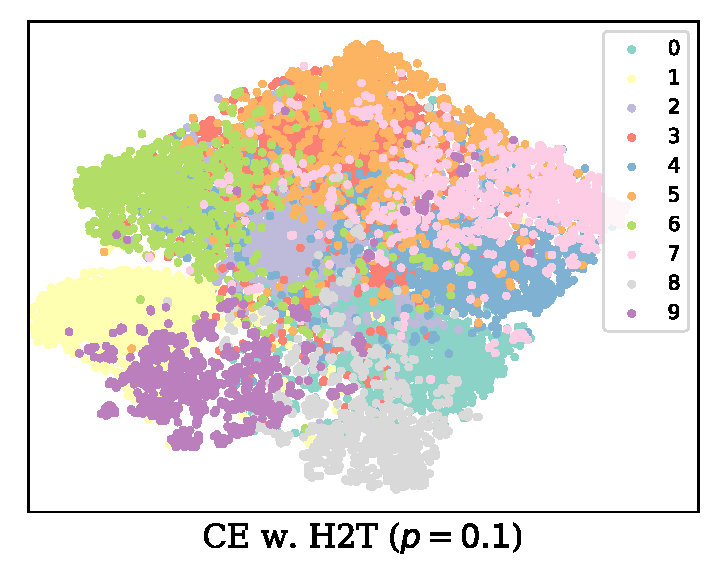}
        \includegraphics[width=0.31\linewidth]{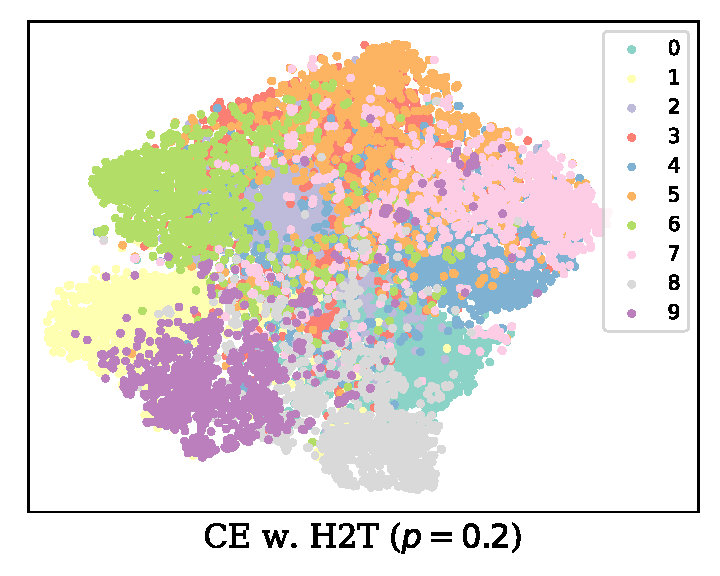}
        \\
        \includegraphics[width=0.31\linewidth]{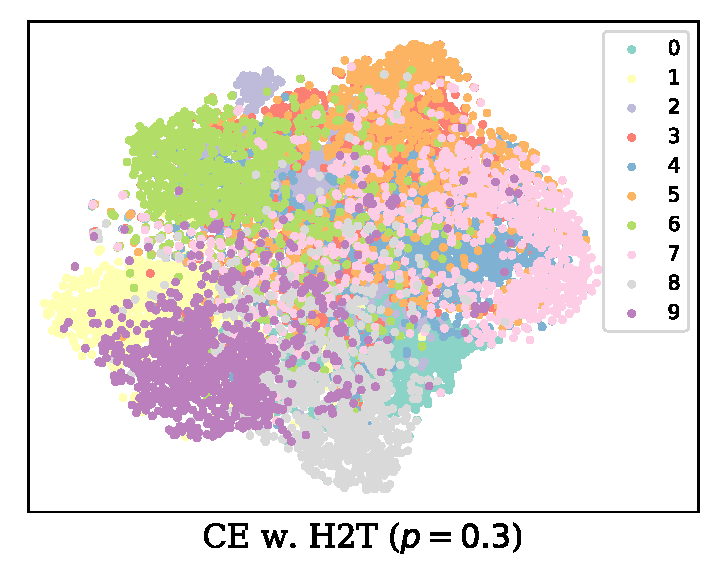}
        \includegraphics[width=0.31\linewidth]{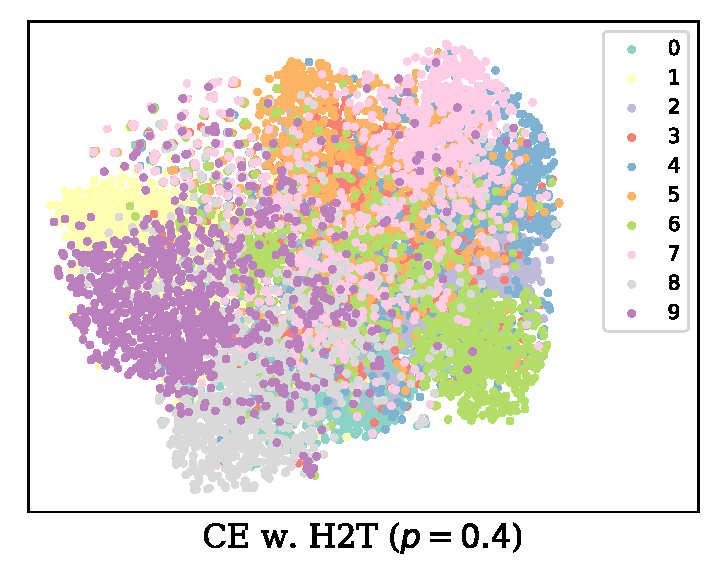}
        \includegraphics[width=0.31\linewidth]{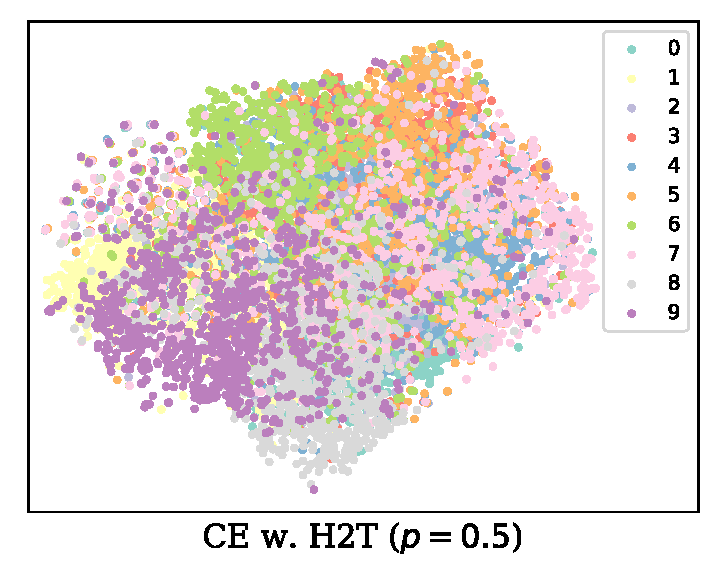}
        \\
        \includegraphics[width=0.31\linewidth]{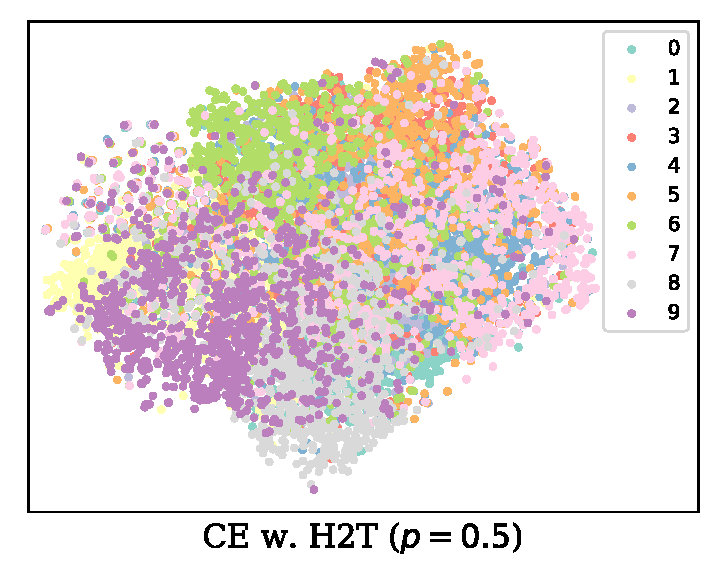}
        \includegraphics[width=0.31\linewidth]{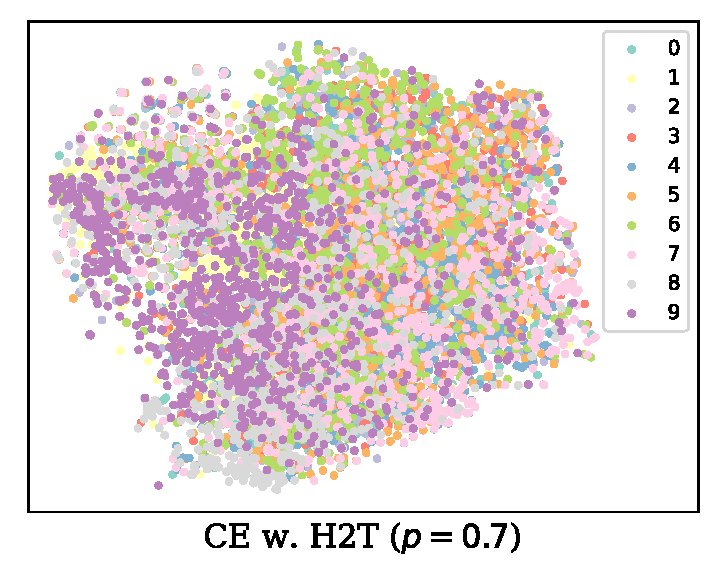}
        \includegraphics[width=0.31\linewidth]{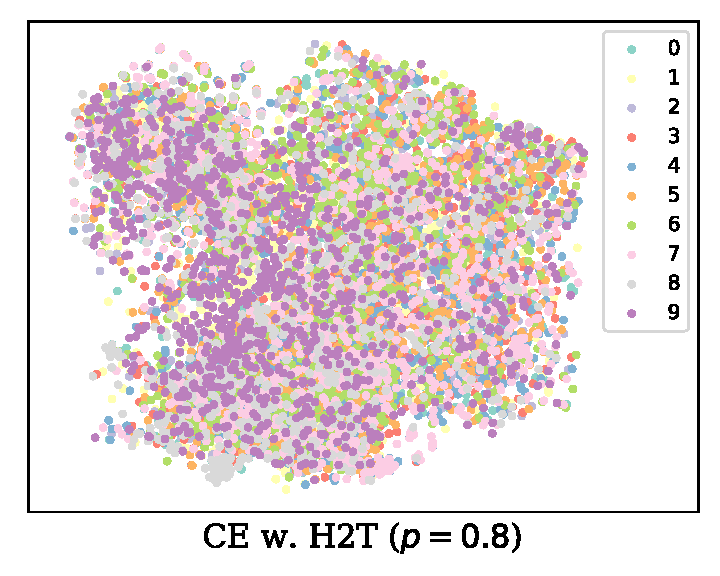}
        \\
        \includegraphics[width=0.31\linewidth]{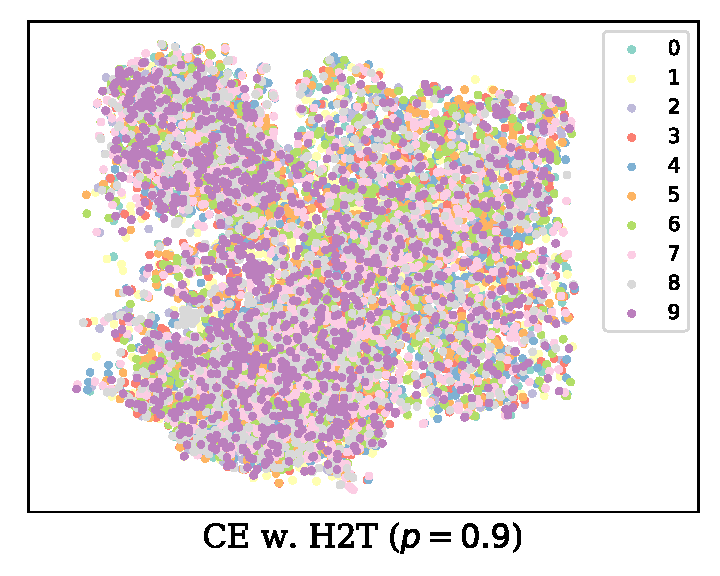}
        \includegraphics[width=0.31\linewidth]{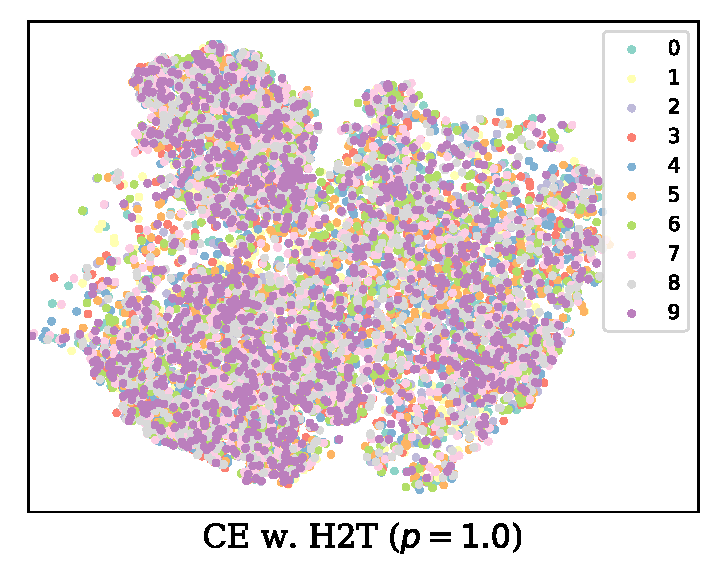}  
\caption{Visualization of feature distribution w.r.t. $p$.}
\label{fig:supp_vis_p}
\end{figure*}

This section includes visualization results for fusion ratio $p$ mentioned in Section~\ref{sec:H2T_method} of the main paper. 
We use CIFAR10-LT for easy visualization.
Figure~\ref{fig:supp_vis_p} shows that as the value of $p$ increases, the interior of each category becomes saturated first, and after $p$ surpasses 0.4, the samples within each category gradually become more dispersed. 
These findings suggest that the selection of an appropriate fusion ratio is crucial in balancing the preservation of the interior samples while ensuring that the model is able to capture the tail samples effectively. 
By gaining a deeper understanding of the relationship between the fusion ratio and the distribution of sample points, we can optimize the performance of the model and ensure its suitability for a range of applications.

(Figure~\ref{fig:supp_vis_p} is located on the next page.)

\section{Comparison with Existing Methods} 

\noindent {\footnotesize$\bullet$} \textit{Clarify H2T}. 
H2T constructs virtual tail features to simulate potential sample variations by grafting part of head features into the tail, enabling tail samples to populate the class boundary, as depicted in Fig.~1. 
It enhances generalization by \textbf{explicitly introducing perturbations}.

\noindent {\footnotesize$\bullet$} \textit{Comparison with influence-balanced loss (IB)~\cite{ParkS2021ICCV}}.
Reducing decision boundary bias is a common goal in long-tail learning.
Different from H2T, IB implicitly addresses this issue from an optimization perspective.
The gradients in stage \uppercase\expandafter{\romannumeral2} of IB is $\lambda_k \cdot \left({\nabla \mathcal{L}_k}/{\|\nabla \mathcal{L}_k\|}\right)$. 
Essentially, IB directly specifies the gradient descent step size to promote a more substantial gradient descent for tail classes to finetune the classifier. 

\noindent {\footnotesize$\bullet$} \noindent\textit{Comparison with CMO~\cite{ParkS2022Majority}}. 
CMO allows the transformation of discrete sample values into continuous ones by mixup inputs and labels using CutMix. 
No new samples are created essentially.
H2T constructs new virtual tail features where labels are not involved in fusion, which can simulate samples that are absent from the training set. 

\noindent {\footnotesize$\bullet$} \noindent\textit{Comparison with head-to-tail network (HTTN)~\cite{xiao2021does}}.
HTTN \textbf{indirectly} utilizes head samples to transfer the prototypes.
It computes the attention ($\alpha_{zj}$) between tail and head prototypes, then uses $\alpha_{zj}$ and the relationship ($W_{transfer}$) between head prototypes and head classifier to identify tail classifier.
In H2T, tail classes \textbf{directly} employ a portion of head features to simulate unseen tail samples and thus populate the class boundary. 

\section{Additional Rational Analysis}
The addition of the inequality obtained from correctly classified tail samples to Eq.~\ref{eq:actual_z} does not eliminate the common terms on both sides of the inequality, and consequently, it does not lead to a meaningful result.
We consider the inequality relationship obtained by correctly classified tail samples and fused head samples:
\begin{equation}\label{eq:actual_z2}
\begin{split}
    \dot{w}_{t}^T\dot{f}_{t}+ \mathcolorbox{pink}{\ddot{w}_{t}^T\ddot{f}_{t}} & > \dot{w}_{h}^T\dot{f}_{t}+\mathcolorbox{CornflowerBlue}{\ddot{w}_{h}^T\ddot{f}_{t}} \rightarrow \text{for correct tail}\\ 
    \dot{w}_{h}^T\dot{f}_{h}+\mathcolorbox{CornflowerBlue}{\ddot{w}_{h}^T\ddot{f}_{t}} & > \dot{w}_{t}^T\dot{f}_{h}+\mathcolorbox{pink}{\ddot{w}_{t}^T\ddot{f}_{t}} \rightarrow \text{for fused head} 
\end{split} \nonumber
\end{equation}
Then, we can derive Eq.~\ref{eq:theta2}, which encourages $w_t$ to be in closer proximity to tail samples, as depicted in Figure~\ref{fig:rationale}.

\end{document}